\documentclass{article}

\usepackage{times}
\usepackage{graphicx} 
\usepackage{subcaption} 
\usepackage[numbers]{natbib}

\usepackage{algorithm}
\usepackage{algorithmic}

\usepackage{bm}
\usepackage{amsmath}
\usepackage{nips_style} 
\usepackage{sidecap}
\usepackage[export]{adjustbox}
\newcommand{\lp}{\left(}
\newcommand{\rp}{\right)}
\newcommand{\lb}{\left[}
\newcommand{\rb}{\right]}

\newcommand{\eps}{\epsilon}
\newcommand{\Keps}{K_{\eps}}

\newcommand{\X}{{\bf X}}

\newcommand{\x}{{\bf x}}
\newcommand{\y}{{\bf y}}
\newcommand{\xp}{\x^{'}}
\newcommand{\xps}{\x^{'(s)}}

\newcommand{\mutheta}{{\bm \mu}_{\thetav}}

\newcommand{\muthetap}{{\bm \mu}_{\thetapv}}

\newcommand{\w}{{\bf w}}
\newcommand{\thetamat}{ {\bm \Theta}}
\newcommand{\thetav}{{\bm \theta}}
\newcommand{\thetapv}{{\bm \theta^{'}}}
\newcommand{\muhattheta}{\hat{{\bm \mu}}_{\thetav}}

\newcommand{\Sigmahattheta}{\hat{{\bm \Sigma}}_{\thetav}}
\newcommand{\muhatthetap}{\hat{{\bm \mu}}_{\thetapv}}

\newcommand{\Sigmahatthetap}{\hat{{\bm \Sigma}}_{\thetapv}}
\newcommand{\Sigmatheta}{{\bm \Sigma}_{\thetav}}

\newcommand{\p}{\pi}
\newcommand{\eye}{{\bm I}}

\newcommand{\muthetapm}{{\bm \mu}^{(m)}_{\thetapv}}
\newcommand{\muthetam}{{\bm \mu}^{(m)}_{\thetav}}

\newcommand{\alpham}{\alpha^{(m)}}

\newcommand{\muthetapj}{\mu_{\thetapv j}}
\newcommand{\muthetaj}{\mu_{\thetav j}}
\newcommand{\muthetapjbar}{{\bar \mu}_{\thetapv j}}
\newcommand{\muthetajbar}{{\bar \mu}_{\thetav j}}

\newcommand{\muthetapmj}{\mu^{(m)}_{\thetapv j}}
\newcommand{\muthetamj}{\mu^{(m)}_{\thetav j}}

\newcommand{\sigmasqthetap}{\sigma^2_{\thetapv j}}
\newcommand{\sigmasqtheta}{\sigma^2_{\thetav j}}
\newcommand{\sigmathetathetap}{\sigma_{\thetav\thetapv j}}
\newcommand{\sigmathetaptheta}{\sigma_{\thetapv\thetav j}}
\newcommand{\thetan}{{\bm \theta}_n}
\newcommand{\xn}{{\bm \x}_n}
\newcommand{\yj}{y_j}
\newcommand{\kthetapN}{{\bm k}_{\thetapv \thetamat j}}
\newcommand{\kthetaN}{{\bm k}_{\thetav \thetamat j}}
\newcommand{\kthetapthetap}{k_{\thetapv \thetapv j}}
\newcommand{\kthetatheta}{k_{\thetav \thetav j}}
\newcommand{\kthetaptheta}{k_{\thetapv \thetav j}}
\newcommand{\kthetathetap}{k_{\thetav \thetapv j}}
\newcommand{\KNN}{{\bm K}_{\thetamat\thetamat j}}
\newcommand{\errorcond}{\mathcal{E}_u(\alpha)}
\newcommand{\erroruncond}{\mathcal{E}(\alpha)}
\DeclareMathOperator\simulator{sim}

\DeclareMathOperator\median{median}

\DeclareMathOperator\expdist{Exp}

\title{GPS-ABC: Gaussian Process Surrogate Approximate Bayesian Computation}

\author{
Edward Meeds \\
Informatics Institute\\
University of Amsterdam\\
\texttt{tmeeds@gmail.com} \\
\And
Max Welling \\
Informatics Institute\\
University of Amsterdam\\
\texttt{welling.max@gmail.com} 
}
\nipsfinalcopy
\begin{document} 
	\vskip -0.3in
\maketitle



%

%
%
\begin{abstract} 
Scientists often express their understanding of the world through a computationally demanding simulation program. Analyzing the posterior distribution of the parameters given observations (the inverse problem) can be extremely challenging. The Approximate Bayesian Computation (ABC) framework is the standard statistical tool to handle these likelihood free problems, but they require a very large number of simulations. In this work we develop two new ABC sampling algorithms that significantly reduce the number of simulations necessary for posterior inference. Both algorithms use confidence estimates for the accept probability in the Metropolis Hastings step to adaptively choose the number of necessary simulations. Our GPS-ABC algorithm stores the information obtained from every simulation in a Gaussian process which acts as a surrogate function for the simulated statistics. Experiments on a challenging realistic biological problem illustrate the potential of these algorithms.
\end{abstract} 
\section{Introduction} \label{introduction}
The morphogenesis of complex biological systems, the birth of neutrons stars, and weather forecasting are all natural phenomena whose understanding relies deeply upon the interaction between simulations of their underlying processes and their naturally observed data.  Hypotheses positing the generation of observations evolve after critical evaluation of the match between simulation and observation.

This hypothesis--simulation--evaluation cycle is the foundation of {\em simulation-based modeling}.  For all but the most trivial phenomena, this cycle is grossly inefficient.  A typical simulator is a complex computer program with a potentially large number of interacting parameters that not only drive the simulation program, but are also often the variables of scientific interest because they play a role in explaining the natural phenomena.  Choosing an optimal value setting for these parameters can be immensely expensive.  

While a single, optimal parameter setting may be useful to scientists, often they are more interested in the {\em distribution} of parameter settings that provide accurate simulation results \cite{schafer2012likelihood, bazin2010likelihood, ratmann2007using}.  The interaction between parameters can provide insight regarding not only the properties of the simulation program, but more importantly, the underlying phenomena of interest.  The main challenges that we address in this paper are 1) simulation-based modeling in a {\em likelihood-free} setting (we do not have a model in the typical machine learning sense, and therefore we do not have a standard likelihood function), and 2) running simulations is very expensive.

The first challenge is partly addressed by the {\em approximate Bayesian computation} (ABC) approach to sampling in likelihood-free scenarios \cite{tavare1997inferring,beaumont2002approximate}.  The ABC approach will be described in a later section, but in short it uses the distance between simulated and observed data as a proxy for the likelihood term in the parameter posterior.  ABC provides the necessary framework to make progress in simulation-based modeling, but it is a very inefficient approach, even on simple problems.

The second challenge is approached with a surrogate model in mind.  This means that every simulation (parameters and result) is stored and used to maintain a surrogate of the mapping from parameters to result.  By carefully constructing an approximate Markov chain Monte Carlo (MCMC) sampler, we are able to sample from the parameter distribution in such a way that our sampling error is controlled and decreases over time.  The main advantage of this approach is that by accepting some bias, we are able to very efficiently sample from the approximate posterior because parameters can sometimes be accepted within MCMC with high confidence by relying on the surrogate and thus avoiding expensive simulations.

In this paper we present a procedure for approximate Bayesian inference using a Gaussian process surrogate for expensive simulation-based models.  In our approach, simulations that are run over the course of the inference procedure are incorporated into a GP model, gradually improving the surrogate over time. Using MCMC with a Metropolis-Hastings (MH) accept/reject rule, our method uses the surrogate and its uncertainty to make all MH decisions. The key insight is that the uncertainty in the acceptance probability is used to determine if simulations are required to confidently proceed to the MH step.  If the uncertainty is high, and we are likely to make an error, then more simulations are run.  


\section{Approximate Bayesian Computation}\label{lfinference}
The current state-of-the-art for simulation-based model inference is {\em likelihood-free} or {\em approximate Bayesian computation} methods \cite{sisson:2010,marin:2012}.  In this section we briefly introduce likelihood-free inference with a focus on MCMC inference procedures as our modeling approach will extend naturally from this work.  

In likelihood-free sampling, we do not have a model in the typical sense.  Instead, we have access to a simulator that generates pseudo-data that, given an accurate model of the world, look like the observations.  The goal is to infer the parameters of the simulator which produce accurate pseudo-data. Importantly, we do not have access to a tractable likelihood function. We now describe in detail the likelihood-free set-up and in particular MCMC sampling techniques.  

One of the primary goals of Bayesian inference is to infer the posterior  distribution of latent variables $\thetav$ using its prior $\pi(\thetav)$ and its data likelihood $\pi(\y | \thetav)$: 
\begin{equation}
	\p( \thetav | \y ) = \frac{\p(\thetav) \p( \y | \thetav)}{\int \p(\thetav) \p( \y | \thetav) d \thetav}
\end{equation}
where $\thetav$ is a vector of latent parameters and $\y$ is the observed data set.  
In simulation-based modeling, we do not have access to the likelihood $\p( \y | \thetav)$. Instead our model of the world consists of a simulator that generates samples  
$\x  \overset{\simulator}{\sim} \pi( \x| \thetav )$ (where we indicate that the simulator was run with parameters $\thetav$ and returns pseudo-data $\x$). Simulation results $\x$ are then compared with the observations $\y$ through a distribution $\pi_\epsilon(\y|\x,\thetav)$, which measures how similar $\x$ is to $\y$. The distribution is parameterized by $\epsilon$ that controls the acceptable discrepancy between $\x$ and $\y$.  We can thus approximately infer the posterior distribution as
\begin{equation}
	\p_\epsilon( \thetav | \y ) = \frac{\p(\thetav)}{\p(\y)} \int \pi_\epsilon(\y|\x) \p(\x|\thetav) d \x \label{eq:lfmarginal}
\end{equation} 
This approximate posterior is only equal to the true posterior for $\pi_{\epsilon=0}(\y|\x)=\delta (\y,\x)$ where $\delta(\cdot)$ is the delta-function. For the exact posterior, one could apply a rejection sampling procedure that would repeatedly sample $\thetav \sim \p(\thetav)$, run a simulation $\x \overset{\simulator}{\sim} \pi( \x| \thetav )$, then accept $\thetav$ only if it equals $\y$. For continuous data, $\epsilon$ acts as a slack variable because equality cannot be achieved.  However, we prefer small $\epsilon$ because this will improve our approximation to the true posterior.  Unfortunately, there remains an unavoidable trade-off between approximation bias (large for large $\epsilon$) and rejection rate (large for small $\epsilon$).  
 
\subsection{Marginal and Pseudo-Marginal ABC}\label{sec:abcmcmcsamplers}
Instead of the rejection sampler described in the previous section (which is hopeless in high dimensions), we now describe two MCMC procedures, the marginal sampler and the pseudo-marginal sampler \cite{andrieu2009pseudo}. At every iteration of MCMC we propose a new $\thetapv \sim q(\thetapv | \thetav)$. Next we sample $S$ samples $\xp_s\sim \pi( \xp | \thetapv ), ~s=1..S$ from a simulator. From these samples we approximate the marginal likelihood as follows,
\begin{equation}
 \p_\epsilon(\y|\thetapv) = \int \pi_\epsilon(\y|\x) \p(\x|\thetav) d \x \approx \frac{1}{S} \sum_{s=1}^S \pi_{\epsilon}(\y | \x^{(s)}, \thetav )
 \label{eq:marginal-approx}
\end{equation}
We accept the proposed parameter $\thetapv$ with probability equal to,
\begin{equation}
	\alpha( \thetapv | \thetav )  = \min \lp 1,  \frac{\pi(\thetapv) \sum_s \pi_{\epsilon}(\y | \xps, \thetapv )q( \thetav | \thetapv) }{\pi(\thetav) \sum_s \pi_{\epsilon}(\y | \x^{(s)}, \thetav )q( \thetapv | \thetav) } \rp
\end{equation}
where the estimate of the marginal likelihood in the denominator (based on $\{\x_s,\thetav\}$) is carried over from the previous iteration. It can be shown that this algorithm is an instance of the more general pseudo-marginal procedure \cite{andrieu2009pseudo} because the estimate of the marginal likelihood is unbiased. From this we can immediately infer that this Markov chain converges to the posterior $\p_\epsilon(\thetav|\y)$. Interestingly, there is an alternative view of this sampler \cite{sisson:2008} that interprets it as a Markov chain over the extended state $\{\thetav,\x_1,...,\x_S\}$ which also leads to the conclusion that the samples $\thetapv$ will asymptotically follow the distribution $\p_\epsilon(\thetav|\y)$. 

Unfortunately, it is well known that pseudo-marginal samplers can suffer from slow mixing. In particular, when through a ``lucky'' draw our marginal likelihood estimate in Eqn. \ref{eq:marginal-approx} attains a large value, then it is very difficult for the sampler to mix away from that state. To avoid this behavior it is sometimes beneficial to re-estimate the denominator (as well as the numerator) in every iteration. This procedure is more expensive and does not guarantee convergence to $\p_\epsilon(\thetav|\y)$ (unless $S\rightarrow\infty$), but can result in much better mixing. We will call this the marginal LF MCMC method \cite{andrieu2009pseudo}.

While for the pseudo-marginal approach we can interpret the fluctuations induced by estimating the $\p(\y|\thetav)$ from a finite sample set as part of randomly proposing a new state, this is no longer true for the approximate marginal MCMC. For the latter it is instructive to study the uncertainty in the acceptance probability $\alpha( \thetapv | \thetav )$ due to these fluctuations: repeatedly estimating $\p(\y|\thetav)$ with $S$ samples will produce a distribution over $\alpha$.  Clearly for small $S$ the distribution will be wider while for very large $S$ it will approach a delta peak.  Our approach uses this distribution directly to determine the confidence in the MH accept step, allowing it to implicitly set $S$ based on the local uncertainty.  This will be further discussed in next sections.

Besides the marginal and pseudo-marginal approaches to ABC, there is a large body of work using sequential Monte Carlo sampling to approach this problem \cite{sisson2007sequential,beaumont2009adaptive,toni2009approximate}.

\section{The Synthetic Likelihood}\label{populationmodel}

We next discuss the approximation introduced by Wood \cite{wood2010statistical}, who models the simulated pseudo-data $\{\x_1,..,\x_S\}$ at parameter value $\thetav$ using a normal distribution, i.e. $\pi(\x|\theta) \approx \mathcal{N}(\x| \muhattheta, \Sigmahattheta)$. 
We will later replace this with a Gaussian process. 
The procedure is very simple: we draw $S$ samples from our simulator and compute first and second order statistics,
\begin{eqnarray}
	\muhattheta    & = & \frac{1}{S} \sum_s \x^{(s)} \label{eq:muhattheta}\\
	\Sigmahattheta & = & \frac{1}{S-1} \sum_s \lp \x^{(s)} - \muhattheta \rp \lp \x^{(s)} - \muhattheta \rp^T \label{eq:Sigmahattheta}
\end{eqnarray}
Using  estimators $\muhattheta$ and $\Sigmahattheta$ we set $\pi(\x|\theta) = \mathcal{N}\lp \muhattheta,\Sigmahattheta\rp$. If we moreover also use a Gaussian kernel,
\begin{equation}
\pi_{\eps}(\y|\x) = \Keps\lp \y, \x \rp = \frac{1}{(2\pi\eps)^{J/2}} e^{-\frac{1}{2\epsilon^2}\lp\x-\y\rp^T \lp\x-\y\rp }
\end{equation}
where $J$ is the dimension of $\y$, we can then analytically integrate over $\x$ in Eqn~\ref{eq:lfmarginal} giving the 
{\em synthetic-ABC} likelihood:
\begin{eqnarray}
	\pi(\y|\thetav) & = & \int \Keps\lp \y, \x \rp \mathcal{N}\lp \muhattheta,\Sigmahattheta\rp d\x \\
	& = & \mathcal{N}\lp \muhattheta,\Sigmahattheta + \epsilon^2\eye \rp
\end{eqnarray}
which has the satisfying result that likelihood of $\y | \thetav$ is the density under a Gaussian model at each  simulation parameter setting $\thetav$.

This approximation has two advantages. First, we can take the limit $\epsilon \rightarrow 0$. This implies that the bias introduced by the need to use a distribution $\p_{\eps}(\y|\thetav)$ to measure the similarity between simulations $\x$ and observations $\y$ is now removed. But this is traded off with the bias introduced by  modeling the simulations from $\p(\x|\thetav)$ with a normal distribution. The second advantage was the main motivation in \cite{wood2010statistical}, namely that this procedure is more robust for extremely irregular probability distributions as encountered in chaotic or near chaotic simulation dynamics. 

A marginal sampler based on a Gaussian approximation (Algorithm~\ref{algo:mh}) has the following acceptance probability:
\begin{equation}
	\alpha( \thetapv | \thetav ) =  \min \lp 1,  \frac{\pi(\thetapv)\mathcal{N}\lp \muhatthetap,\Sigmahatthetap + \epsilon^2 \eye\rp q( \thetav | \thetapv) }{\pi(\thetav) \mathcal{N}\lp \muhattheta,\Sigmahattheta + \epsilon^2\eye \rp q( \thetapv | \thetav) } \rp \label{eq:nonrandompopacceptance}
\end{equation}
As with the marginal sampler of Section~\ref{lfinference}, the fact that we estimate first and second order statistics from a finite sample set introduces uncertainty in the accept probability $\alpha( \thetapv | \thetav )$: another run of $S$ simulations would have resulted in different values for these statistics and hence of the accept probability. See Figure~\ref{fig:acceptance_distribution} for an illustration. In the following section we will analyze the distribution over $\alpha( \thetapv | \thetav )$ and develop a method to decide how many simulations $S$ we need in order to be sufficiently confident that we are making the correct accept/reject decision.  Random acceptance probability distributions have been studied in general \cite{nicholls:2012} and for the specific case of Gaussian log-energy estimates \cite{ceperley:1999}.  
\begin{algorithm}[t] 
	\caption{Synthetic-likelihood ABC MH step}
	\label{algo:mh}
	\begin{algorithmic}
		\STATE {\bfseries inputs:} $q, \thetav, \pi(\x|\thetav), S, \eps, \y$ 
		\STATE $\thetapv \sim q(\thetapv | \thetav )$
			\FOR{$s = 1 : S$} 
				\STATE $\xps \overset{\simulator}{\sim} \pi( \x| \thetapv )$,  $\x^{(s)} \overset{\simulator}{\sim} \pi( \x| \thetav )$
			\ENDFOR
			\STATE Set $\muhatthetap$,$\Sigmahatthetap$, $\muhattheta$, $\Sigmahattheta$ using Eqns \ref{eq:muhattheta} and \ref{eq:Sigmahattheta}.
			
			\STATE Set $\alpha$ using Eqn \ref{eq:nonrandompopacceptance}
			\IF{$\mathcal{U}(0,1) \leq \alpha$} 
				\STATE \textbf{return} $\thetapv$
			\ENDIF
			\STATE \textbf{return} $\thetav$
	\end{algorithmic}
\end{algorithm}

\subsection{MCMC with a Random Acceptance Probability}

We now make explicit the role of randomness in the MCMC sampler with synthetic (normal) likelihoods.  At each iteration of the MCMC sampler, we compute estimators $\{ \muhattheta, \Sigmahattheta, \muhatthetap, \Sigmahatthetap\}$ as before using Eqns \ref{eq:muhattheta} and \ref{eq:Sigmahattheta}. To estimate the distribution over accept probabilities we would need $M$ sets of $S$ simulations, which would be too expensive. Instead, we use our Gaussian assumption to derive that the variance of the mean is $1/S$ times the variance in the sample $\{\x_1,...,\x_S\}$,
\begin{equation}
\mutheta \sim \mathcal{N}\left(\muhattheta,\Sigmahattheta/S\right)
\label{eq:meanposterior}
\end{equation}
and similarly for $\muthetap$. This shortcut is important because it allows us to avoid a significant number of expensive simulations and replace them with samples from a normal distribution. 

Given our $M$ samples of $(\mutheta,\muthetap)$, we can compute $M$ samples of $\alpha( \thetapv | \thetav )$ by inserting them into the expression for the randomized MH accept probability:
\begin{equation}
		\alpham  =  \min \lp 1,  \frac{\pi(\thetapv)\mathcal{N}\lp \y | \muthetapm, \Sigmahatthetap + \epsilon^2\eye \rp q( \thetav | \thetapv) }{\pi(\thetav) \mathcal{N}\lp \y | \muthetam, \Sigmahattheta + \epsilon^2\eye \rp q( \thetapv | \thetav) } \rp  \label{eq:montecarloacceptanceprobability}
\end{equation}
We now derive a procedure to estimate the probability of making an error in an accept/reject decision and a threshold $\tau$ for actually making the decision.  The error of making an incorrect decision can either be measured conditioned on $u \sim \mathcal{U}(0,1)$ (the uniformly distributed draw used in the MH decision), or unconditionally, by integrating over $\mathcal{U}(0,1)$.  First we start with the conditional error which trivially extends to the unconditional error by averaging.  

\begin{figure}[t]
\vskip 0.2in
\begin{center}
\centerline{\includegraphics[scale=0.35]{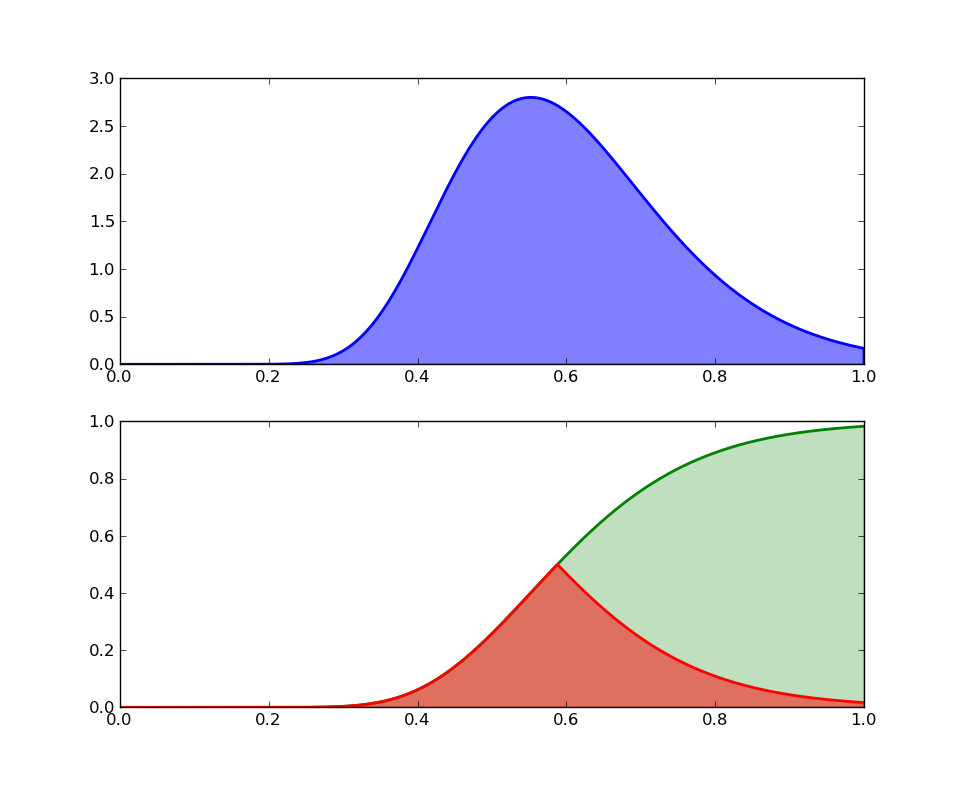}}
\vskip -0.2in
\caption{\small{An example of $p(\alpha)$, the distribution over acceptance probabilities ({\bf top}) and  its CDF shown folded at its median ({\bf bottom}).}}
\label{fig:acceptance_distribution}
\end{center}
\vskip -0.2in
\end{figure} 

If $u \leq \tau$, then we accept the MH proposal and move to the proposed state.  The probability of making an error in this case is $P( \alpha < u )$ (i.e. the probability we should actually reject):
\begin{equation}
  P( \alpha < u ) = \frac{1}{M} \sum_m \lb \alpham < u\rb
\end{equation}
Similarly, if $u > \tau$ then we reject, and the error is $P( \alpha > u )$ (i.e. the probability we should actually accept):
\begin{equation}
  P( \alpha > u ) = \frac{1}{M} \sum_m \lb \alpham \geq u\rb 
\end{equation}
The total conditional error is therefore:
\begin{equation}
	\errorcond = \lb u \leq \tau\rb P( \alpha < u ) + \lb u > \tau\rb P( \alpha \geq u ) \label{eq:conditionalerror}
\end{equation}
and the total unconditional error is:
\begin{equation}
	\erroruncond = \int \errorcond \mathcal{U}(0,1) d u \label{eq:unconditionalerror}
\end{equation}
which can again be estimated by Monte Carlo or grid values of $u$.  The analytic calculation of $\erroruncond$ is the area under the cumulative distribution function of $p(\alpha)$ {\em folded} at $\tau$ (see Figure~\ref{fig:acceptance_distribution}).  This integral is also known as the {\em mean absolute deviation} \cite{xue:2011} which is minimized at the median of $p(\alpha)$ (the value of $\alpha$ where the CDF equals $1/2$), justifying our decision threshold $\tau = \median( \alpha )$ (also determined by samples $\alpham$).

With this in hand, we now have the necessary tools to construct an adaptive synthetic-likelihood MCMC algorithm that uses $\erroruncond$ as a guide for running simulations (Algorithm~\ref{algo:randomizedpopmh}).  At the start of each MH step, $S_0$ simulations are run for both $\thetav$ and $\thetapv$; estimators are computed; then $M$ $\alpham$ are sampled.  Based on these samples, the median and $\erroruncond$ is computed.  Note that this phase of the algorithm is very cheap; here we are sampling from $J$ bivariate Gaussian distributions to compute Monte Carlo estimates of $\tau$ and $\erroruncond$, so $M$ can be set high without a computational hit, though in practice $M<100$ should be fine.  While $\erroruncond > \xi$, $\Delta S$ new simulations are run and the estimators updated, along with new draws of $\alpham$, etc.  The user-defined error threshold $\xi$ is a knob which controls both the accuracy and computational cost of the MCMC.  New simulations can be run at either $\thetav$ or $\thetapv$; we run simulations at both locations, though selecting one over the other  based on the higher individual mean uncertainty could result in fewer simulations. As $S$ increases, the uncertainty around $p(\alpha)$ decreases such that $\erroruncond < \xi$; once this occurs, the MH is now {\em confident} and it proceeds using the usual acceptance test, with $\tau$ as the acceptance threshold.  
 
In many cases, the actual number of simulations required at each step can be quite small, for example when one parameter setting is clearly better than an another (where the median is at or close to $0$ or $1$).  Nevertheless, there remains a serious drawback to this algorithm for expensive simulations: all simulations are discarded after each MH step; a great waste considering the result is a single binary decision.  Using a Gaussian process surrogate, described in the next section, we will remember all simulations and use them to gradually improve the surrogate and as a consequence, eventually eliminate the need to run simulations.
\begin{algorithm}[t]
	\caption{Adaptive Synthetic-likelihood ABC MH step}
	\label{algo:randomizedpopmh}
	\begin{algorithmic}
		\STATE {\bfseries inputs:} $q, \thetav, \pi( \x| \thetav ), S_0, \Delta S, \eps, \y, \xi$ 
		\STATE $\thetapv \sim q(\thetapv | \thetav )$
		\STATE Init $c=1$, $S=S_0$
		\REPEAT
			\FOR{$s = c : c+S$} 
				\STATE $\xps \overset{\simulator}{\sim} \pi( \x| \thetapv )$,  $\x^{(s)} \overset{\simulator}{\sim} \pi( \x| \thetav )$
			\ENDFOR
			\STATE Update $c=S$, $S=S+\Delta S$
			\STATE Set $\muhatthetap$,$\Sigmahatthetap$, $\muhattheta$, $\Sigmahattheta$ using Eqns~\ref{eq:muhattheta} and \ref{eq:Sigmahattheta}.
			\FOR{$m = 1 : M$}
			  \STATE Sample $\muthetapm$, $\muthetam$ using Eqn~\ref{eq:meanposterior}
				\STATE Set $\alpham$ using Eqn~\ref{eq:montecarloacceptanceprobability}
			\ENDFOR
			\STATE Set $\tau = \median(\alpham)$ 
			\STATE Set $\erroruncond$ using Eqn~\ref{eq:unconditionalerror}

			\UNTIL $\erroruncond < \xi$
			\IF{$\mathcal{U}(0,1) \leq \tau$} 
				\STATE \textbf{return} $\thetapv$
			\ENDIF
			\STATE \textbf{return} $\thetav$
	\end{algorithmic}
\end{algorithm}

\section{Gaussian Process Surrogate ABC}\label{skabcgp}

As mentioned in the introduction, in many scientific disciplines simulations can be extremely expensive. The algorithms up till now all have the downside that at each MCMC update a minimum number of simulations need to be conducted. This seems unavoidable unless we store the information of previous simulations and use them to make accept/reject decisions in the future. In particular, we can \emph{learn} the mean and covariance $\mutheta,\Sigmatheta$ of the synthetic likelihood as a function of $\thetav$ and as such avoid having to perform simulations to compute them. There is a very natural tool that provides exactly this functionality, namely the Gaussian process (GP). For our purposes, the GP will ``store'' the simulation runs $\thetav_t,\x_t$ for all simulations conducted during the MCMC run. 
We will use the GP as a ``surrogate'' function for the simulated statistics from which will be able to estimate the marginal likelihood values away from regions where actual simulations were run. Importantly, the GP provides us with uncertainty estimates of the marginal likelihood which will inform us of the need to conduct additional experiments in order to make confident accept/reject decisions. Going from the synthetic likelihood model to the GP represents a change from frequentist statistics in favor of (nonparametric) Bayesian statistics. 
Gaussian processes have recently also become a popular tool in the machine learning literature as surrogates of expensive regression surfaces, such as log-likelihoods \cite{rasmussen:2003}; optimization surfaces \cite{brochu:2010,snoek:2012}; and simulations of physical systems \cite{bilionis2013}.

Our surrogate model and algorithm follow directly from the synthetic-ABC approximation and randomized acceptance algorithm.  The main difference between the two is that in this paper, we model the $J$ statistics as $J$ independent Gaussian processes (recall $J$ is the dimensionality of $\y$). We note that it would be better to model the $J$ statistics using a single joint Gaussian process. This can be done using ``co-Kriging'' or variants thereof  \cite{higdon:2002,boyle:2005} (see also Section~\ref{discussion}).  Although the independence assumption may lead to overconfidence (because it is assuming--falsely--independent evidence), it is also more robust in high-dimensions where the estimator of the full output covariance has high variance (it overfits).  It may be that the mentioned multi-output GPs can provide an appropriate solution by tying the covariance structure across parameter space using a small number of kernel hyperparameters. For the current experiments we found that independent GPs worked well enough to illustrate the algorithm's potential. 

For each $j$, we have the following bivariate conditional distribution:
\begin{equation}
    \begin{bmatrix}
        \muthetapj \\ \muthetaj 
    \end{bmatrix}
		\sim
		\mathcal{N}\lp
    \begin{bmatrix}
        \muthetapjbar \\ \muthetajbar
    \end{bmatrix}
		,
    \begin{bmatrix}
        \sigmasqthetap & \sigmathetaptheta \\ \sigmathetathetap & \sigmasqtheta 
    \end{bmatrix}
		\rp \label{eq:gpposterior}
\end{equation}
where the mean and covariance are determined by the set of $N$ training inputs $\{ \thetan \}$ and $N$ training outputs $\{ \xn \}$ (using only statistic $j$). They are given by the usual expressions for the GP mean and covariance,
\begin{equation}
  \begin{bmatrix}
      \muthetapjbar \\ \muthetapjbar
  \end{bmatrix}
	=
  \begin{bmatrix}
      \kthetapN \\ \kthetaN
  \end{bmatrix}
	\lb
	\KNN + \sigma^2_j \eye
	\rb^{-1}
	\X[:,j]
\end{equation}
and
\begin{equation}
    \hspace{-0.7cm}\begin{bmatrix}
      \sigmasqthetap & \sigmathetaptheta \\ \sigmathetathetap & \sigmasqtheta 
  \end{bmatrix} =  
  \begin{bmatrix}
      \kthetapthetap & \kthetaptheta \\ \kthetathetap & \kthetatheta
  \end{bmatrix}
  -
  \begin{bmatrix}
      \kthetapN \\ \kthetaN
  \end{bmatrix}
	\lb
	\KNN + \sigma^2_j \eye
	\rb^{-1}
  \begin{bmatrix}
      \kthetapN \\ \kthetaN
  \end{bmatrix}^T\nonumber
\end{equation}
where $\kthetaN$ is a $1$ by $N$ vector of kernel evaluations for the $j$'th Gaussian process between $\thetav$ and the input training set $\thetamat$, $\KNN$ is the $j$th kernel matrix evaluated on the training data; $\sigma^2_j$ is the data noise term for the $j$'th statistic (used below in the acceptance ratio), $\X$ is the $N$ by $J$ training output data set and $\X[:,j]$ means selecting column $j$ from the training data, and $\kthetaptheta$ is a single kernel evaluation at $\thetapv$, $\thetav$ for Gaussian process $j$, and similarly for $\kthetapthetap$, $\kthetathetap$, $\kthetatheta$.

\begin{algorithm}[t]
	\caption{GPS-ABC MH step}
	\label{algo:gpmh}
	\begin{algorithmic}
		\STATE {\bfseries inputs:} $q, \thetav, \pi( \x| \thetav ), S_0, \Delta S, \eps, \y, \xi$ 
		\STATE $\thetapv \sim q(\thetapv | \thetav )$
		\REPEAT
			\FOR{$m = 1 : M$}
				\STATE Sample $\muthetapmj$, $\muthetamj$ using Eqn~\ref{eq:gpposterior}
				\STATE Set $\alpham$ using Eqn~\ref{eq:gpacceptanceprobability}
			\ENDFOR
			\STATE Set $\tau = \median(\alpham)$ 
			\STATE Set $\erroruncond$ using Eqn~\ref{eq:unconditionalerror}
			
			\IF{ $\erroruncond > \xi$}
				\STATE Acquire new training point.
			\ENDIF
			\UNTIL $\erroruncond < \xi$
			\IF{$\mathcal{U}(0,1) \leq \tau$} 
				\STATE \textbf{return} $\thetapv$
			\ENDIF
			\STATE \textbf{return} $\thetav$
	\end{algorithmic}
\end{algorithm}

The GPS-ABC algorithm is now run as follows (see Algorithm~\ref{algo:gpmh}). At each MH step, using Eqn~\ref{eq:gpposterior},  for each $j$,  $M$ independent draws of $\muthetap$ and $\mutheta$ are sampled from their bivariate Gaussian process posterior. Note that this is in fact different from the synthetic-ABC model in two ways 1) there are no default simulations at each MH step; instead, the current surrogate model is used to predict both the expectation and uncertainty in simulation output and 2) the Gaussian process takes into account the correlations between $\thetav$ and $\thetapv$. As before, Monte Carlo statistics are computed from acceptance probabilities $\alpham$ as follows,
\begin{equation}
		\alpham  =
		\min \lp 1,  \frac{\pi(\thetapv) \prod_j \mathcal{N}\lp \yj | \muthetapmj, \sigma^2_j + \epsilon^2 \rp q( \thetav | \thetapv) }{\pi(\thetav) \prod_j \mathcal{N}\lp \yj | \muthetamj, \sigma^2_j + \epsilon^2 \rp q( \thetapv | \thetav) } \rp   \label{eq:gpacceptanceprobability}  
\end{equation} 
The error $\erroruncond$ and acceptance threshold $\tau$ are computed from the $M$ samples; if 
$\erroruncond > \xi$, then a procedure for acquiring training points (i.e. simulations) is run, with the objective of reducing uncertainty for this specific MH step.  Again, as with the adaptive synthetic likelihood algorithm, computing the $M$ samples is very cheap.  The procedure is then free to select any input location to run a simulation (whereas before we were forced to run at either $\thetav$ and $\thetapv$), though the new simulation should be impactful for the current MH step.  This means that we can choose locations other than $\thetav$ and $\thetapv$, perhaps trying to limit the number of future simulation runs required in the vicinity.  Analogous to acquisition functions for Bayesian optimization \cite{brochu:2010}, actively acquiring points has the implicit goals of speeding up MCMC, reducing MCMC error, and limiting simulations. Once a new training point is selected and run, the training input-output pair is added to all $J$ Gaussian processes and the model hyperparameters may or may not be modified (with a small number of optimization steps). 

The key advantage of GPS-ABC is that with increasing frequency, we will not have to do any expensive simulations whatsoever during a MH step because the GP surrogate is sufficiently confident about the marginal likelihood surface in that region of parameter space. 

\subsection{Theoretical Aspects of GPS-ABC}\label{sec:theory}
Two of the main contributions of GPS-ABC are {\em MCMC under uncertainty} and the introduction of {\em memory} into the Markov chain; we consider these steps as the only way to reduce the number of expensive simulations and as such a necessary ingredient to GPS-ABC.  Nevertheless, they present major differences from typical Bayesian inference procedures.  

We now address two major theoretical aspects of the GPS-ABC algorithm: the {\em approximate} and {\em adaptive} nature of GPS-ABC.  Although we have postponed formal proofs for future work we have outlined their main arguments below.   

GPS-ABC is {\em approximate} because at each MH-step there is some probability that the chain will make an error, and that this corresponds to an error in the stationary distribution of the Markov chain (i.e. it is an approximation to the stationary distribution).  In \cite{korrattikara:2014}, another approximate MCMC algorithm is presented and it provides a proof  for  an upper bound on the error in the stationary distribution.  The main argument is that if the MH-step error is small and bounded (along with a few other mild conditions), then the error in stationary distribution is bounded as well.  We feel GPS-ABC fits nicely into this same proof framework. 

GPS-ABC is also {\em adaptive} since the approximation to the stationary distribution changes as more training points are added to the Gaussian process (we are learning the surrogate as we run the MCMC).  Two of the major requirements for a valid adaptive MCMC algorithm are {\em diminishing adaptation} and {\em ergodicity} \cite{roberts:2007}.  GPS-ABC satisfies the former as the number of training points acquired over an MCMC rapidly decreases over time.  When the adaptation slows and becomes insignificant, the Markov chain resembles the algorithm in \cite{korrattikara:2014}, which, as we stated above, provides a proof of bounded convergence to the stationary distribution (and hence ergodicity); therefore GPS-ABC satisfies the latter requirement.  


\section{Experiments}
We present three main experiments: 1) a toy Bayesian inference problem, 2) inference of parameters in a chaotic ecological system, and 3) inference of parameters in a stochastic biological system.  The goal of the first experiment is to show the correctness of the adaptive synthetic likelihood algorithm (henceforth ASL-ABC) and GPS-ABC, and to demonstrate the computational improvements achieved by GPS-ABC over both ASL-ABC and Kernel-ABC.  The goal of the second experiment is again to demonstrate computational savings but also illustrate the differences in posterior distributions found with and without the output independence assumption.  Finally, the goal of the third experiment is to demonstrate the potential use of the posterior predictive distribution, along with the parameter posterior distribution, for model selection and comparison in same the spirit as \cite{gelman}.

For ASL-ABC, at each MH-step, there are $S_0$ initial simulations run at both $\thetav$ and $\thetapv$ (akin to the marginal ABC sampler).  Then, while the unconditional MH error is greater than $\xi$, $\Delta S$ additional simulations are run.  Some experiments will vary $S_0$ or $\Delta S$ or both.  By default the full output covariance will be estimated by ASL-ABC, except for one part of experiment 2.  For GPS-ABC, the meaning of $S_0$ is different; it represents the initial number of samples from the prior used to train the GPs.  From that point onward, the GP runs one simulation at a time (while the error is greater than $\xi$).  Although both algorithms permit $\eps > 0$, we have set $\eps = 0$ for all experiments involving ASL-ABC and GPS-ABC.

\subsection{Toy Problem}
As an illustrative example we apply both ASL-ABC and GPS-ABC to estimate the posterior of the rate of an exponential distribution based on the mean statistic from observations from the true generating model with a Gamma prior over the rate parameter;  this is the {\em exponential example} in \cite{turner2012tutorial}.  Let $\w$ be a vector of $N = 500$ draws from an exponential distribution with rate $\thetav^{\star}=0.1$, $w_n \sim \expdist(\thetav^{\star})$;  
then let $\y = \frac{1}{N} \sum w_n$ be the observation statistic. Our simulator is a two-step process: generate $N$ exponential draws with rate $\thetav$; then return their average.  I.e. the same generating process as the observations $\y$.  With a Gamma prior $\pi(\thetav) = \mathcal{G}(\alpha,\beta)$, the goal of this experiment is to infer $p(\thetav | \y)$ using ABC methods.  
The prior distribution has parameters $\alpha = \beta = 0.1$; a very broad prior over $\thetav$.  In this controlled experiment, the same random seed was used for all MCMC runs, which fixes $\y = 9.42$ and $\thetav$-MAP $0.0992$ (not quite $10$ and $0.1$ since they are random draws).
\begin{figure}[ht]
\vskip 0.2in
\begin{center}
\includegraphics[width=0.33\columnwidth]{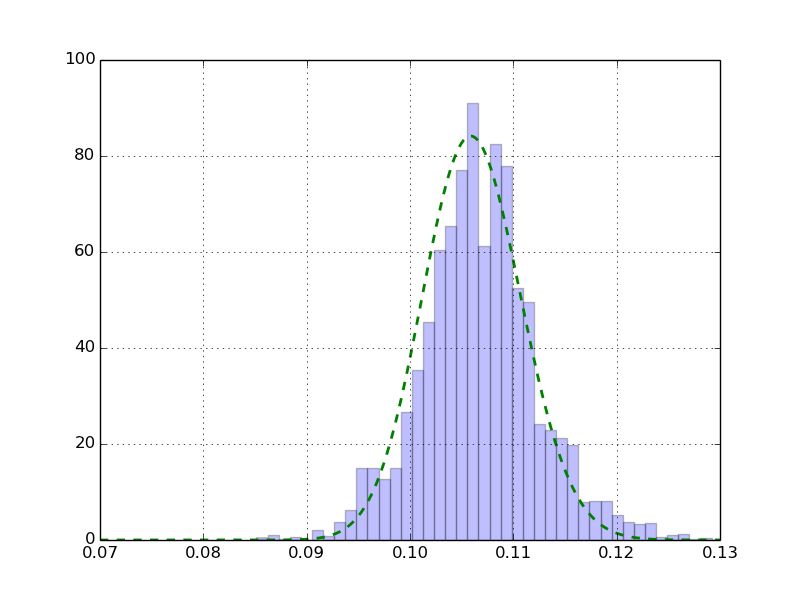}
\hskip -0.1in
\includegraphics[width=0.33\columnwidth]{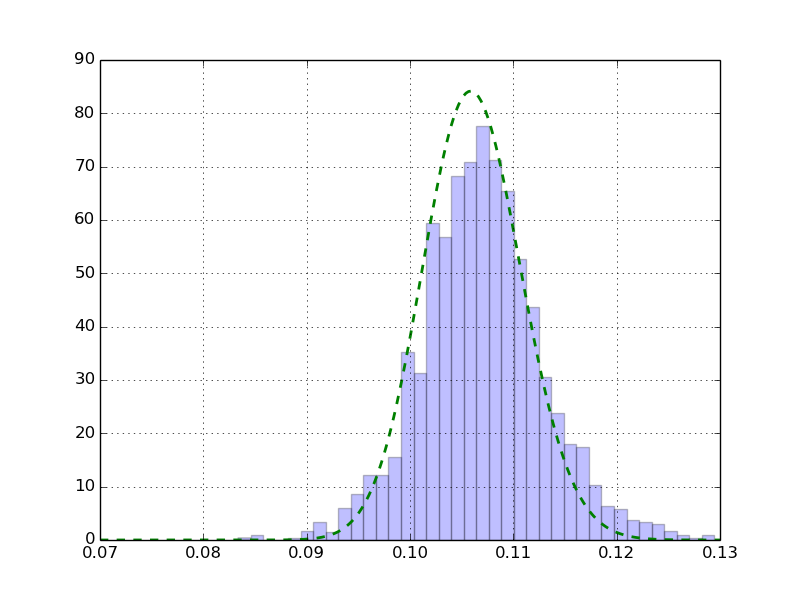}
\hskip -0.1in
\includegraphics[width=0.33\columnwidth]{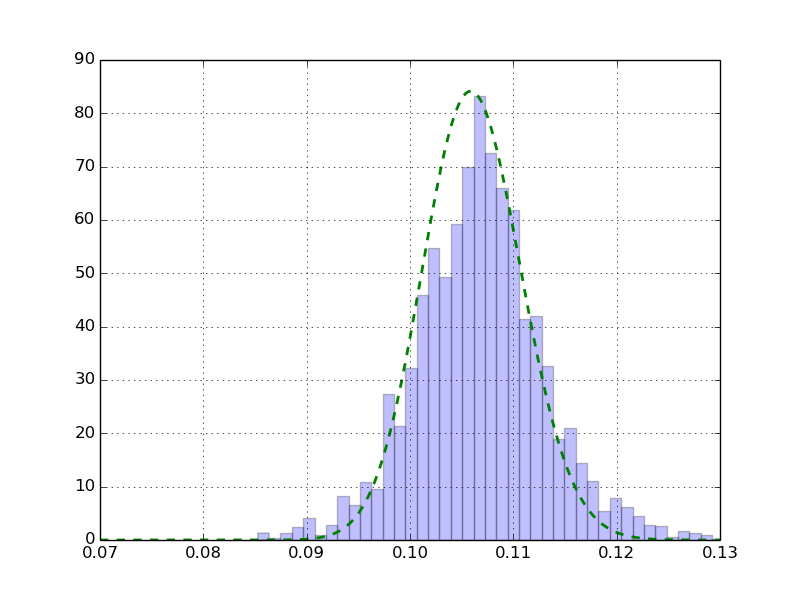}\\
\vskip -0.075in
\includegraphics[width=0.33\columnwidth]{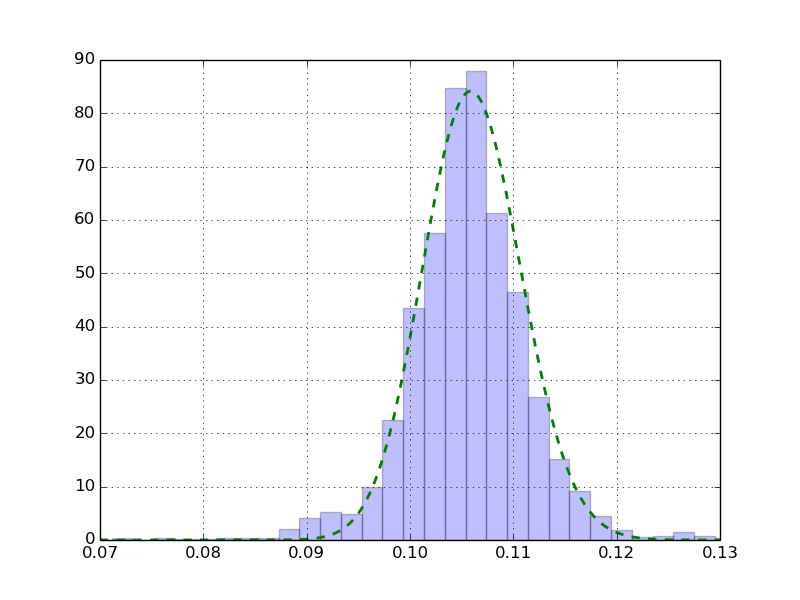}
\hskip -0.1in
\includegraphics[width=0.33\columnwidth]{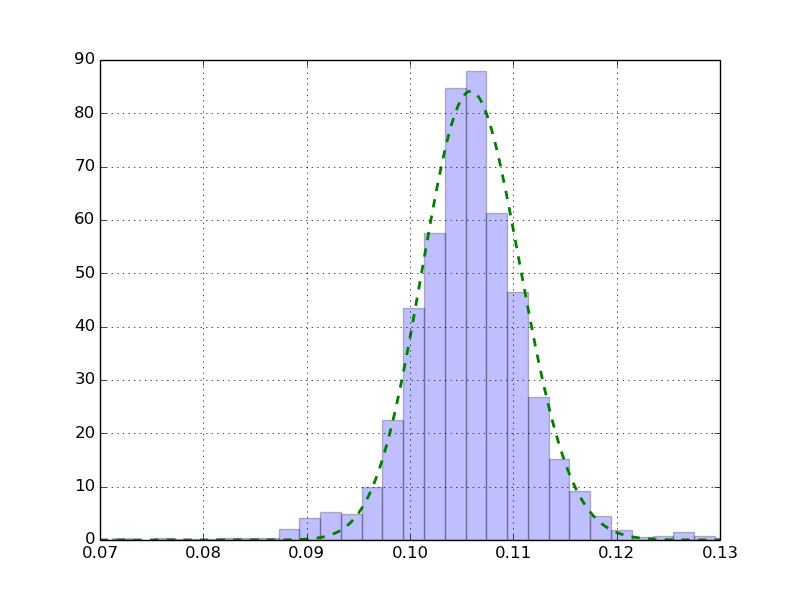}
\hskip -0.1in
\includegraphics[width=0.33\columnwidth]{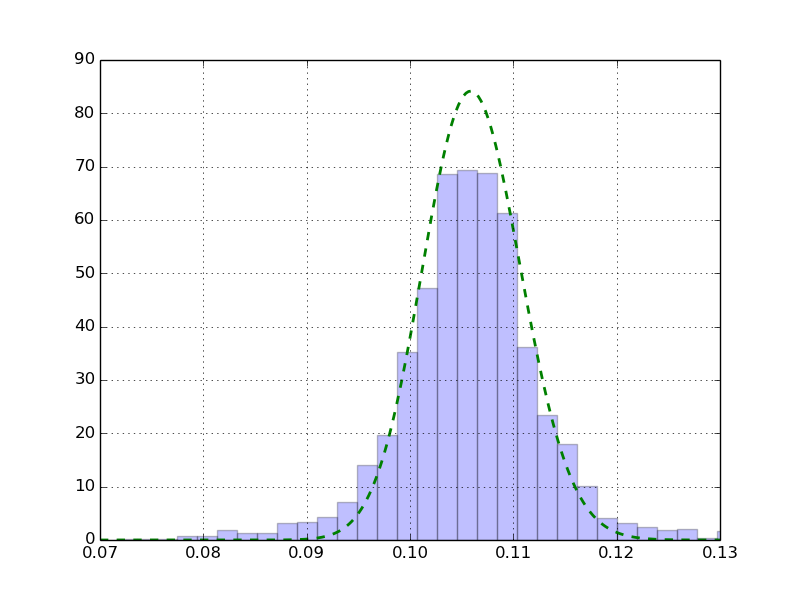}\\
\vskip -0.075in
\includegraphics[width=0.33\columnwidth]{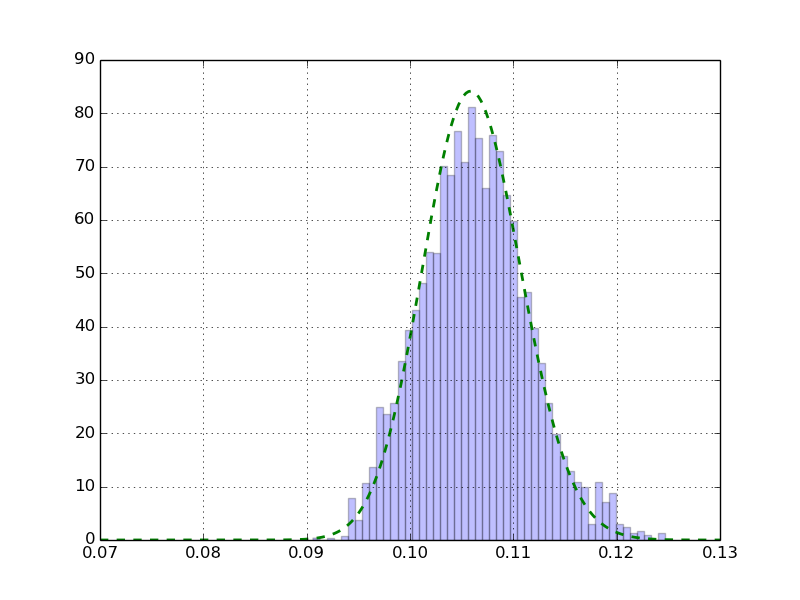}
\hskip -0.1in
\includegraphics[width=0.33\columnwidth]{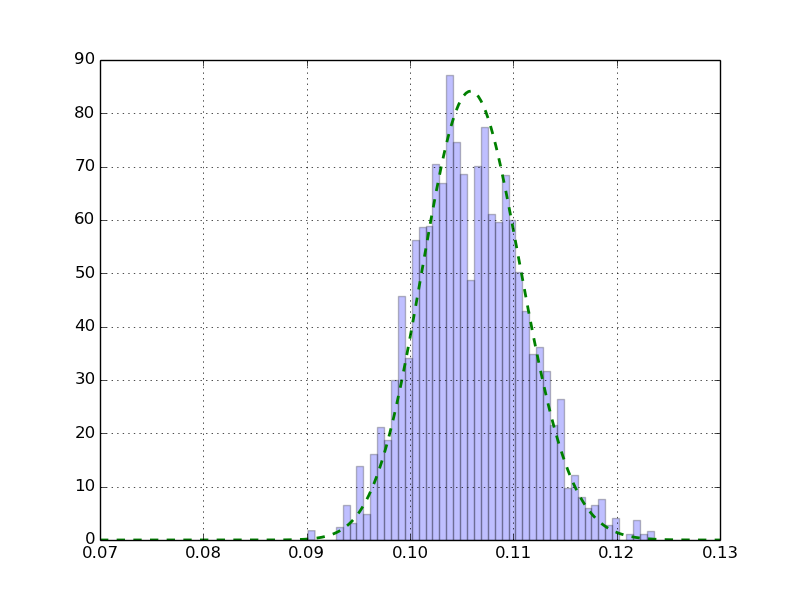}
\hskip -0.1in
\includegraphics[width=0.33\columnwidth]{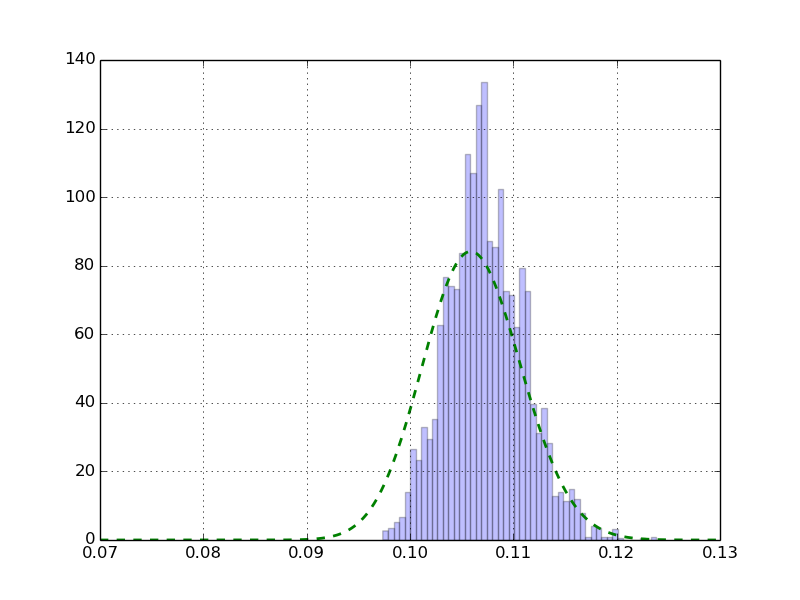}
\caption{\small{Posterior distributions $p(\thetav | \y )$ for the {\em exponential} experiment; the true posterior is shown with the dashed line.  
{\bf Top row}: Kernel-ABC using $\epsilon = \{0.01,0.1,0.5\}$ (note these histograms are based on $50K$ samples, not $10K$, because the histograms for $10K$ were too jagged due to the poor mixing rate of Kernel-ABC). {\bf Middle row}: ASL-ABC using $\epsilon = 0$, $\xi = \{0.05,0.2,0.4\}$ for $S_0 = 5$.  
{\bf Bottom row}: GPS-ABC using $\epsilon = 0$, $\xi = \{0.05,0.2,0.4\}$ for $S_0 = 20$; the number of simulations for these examples were $1297$, $184$, and $29$, respectively.   These are orders of magnitude fewer than the other methods (see Table~\ref{table-exponential} for the number simulation calls).  The slight bias in the bottom-right histogram ($\xi = 0.4$) is an issue with GP hyper-parameter calibration with very few training points ($29$). }}
\label{fig:toycompareposteriors}
\end{center}
\vskip -0.2in
\end{figure} 

In Figure~\ref{fig:toycompareposteriors} we show the posterior distributions produced from $3$ different algorithms: {\em Kernel-ABC}, {\em ASL-ABC}, and {\em GPS-ABC}.  Kernel-ABC uses the marginal-likelihood MCMC sampler (see Section~\ref{sec:abcmcmcsamplers}, with $S=1$), using a Gaussian kernel, $K_{\epsilon}$, and $\epsilon = \{0.01,0.1, 0.2\}$.  At each MH step, values for $\x$ are simulated at both $\thetav$ and $\thetapv$. The true posterior distribution is shown as a dashed line.  All the runs produced $10$K samples, the first $1.5$K of which were discarded.  All algorithms were initialized at $\thetav=1.0$ (very far from the posterior mode).  For ASL-ABC we ran two samplers using $S_0 = \{5,100\}$, $\Delta S = 10$, and $\xi = \{0.4,0.2,0.05\}$; and for GPS-ABC we initialized the GP with $S_0 = 20$ points from the prior, then ran the sampler with $\xi = \{0.4,0.2,0.05\}$.  All algorithms used the same proposal for $\log \thetav$, a normal with $\sigma=0.1$.  

\begin{table}[t]
\caption{Number of simulation calls (in thousands) for the exponential experiment, based on generating $10K$ samples.  Note for Kernel-ABC $\xi$ does not apply, read $\epsilon = \{0.01, 0.1, 0.5\}$}
\label{table-exponential}
\vskip -0.15in
\begin{center}
\begin{small}
\begin{sc}
\begin{tabular}{llccc}
\hline
Model & $S_0$ & $\xi=0.05$ & $\xi=0.2$ & $\xi=0.4$ \\
\hline
ASL-ABC & $5$   & 572  & 135   & 100 \\
ASL-ABC & $100$ & 2197 & 2060  & 2000 \\
GPS-ABC & $20$  & 1.3  & 0.184 & 0.029 \\        
Kernel-ABC   &    N/A       & 20   & 20    & 20 \\
\hline
\end{tabular}
\end{sc}
\end{small}
\end{center}
\vskip -0.1in
\end{table}

\begin{figure}[t]
\vskip -0.1in
\begin{center}
\includegraphics[width=0.33\columnwidth]{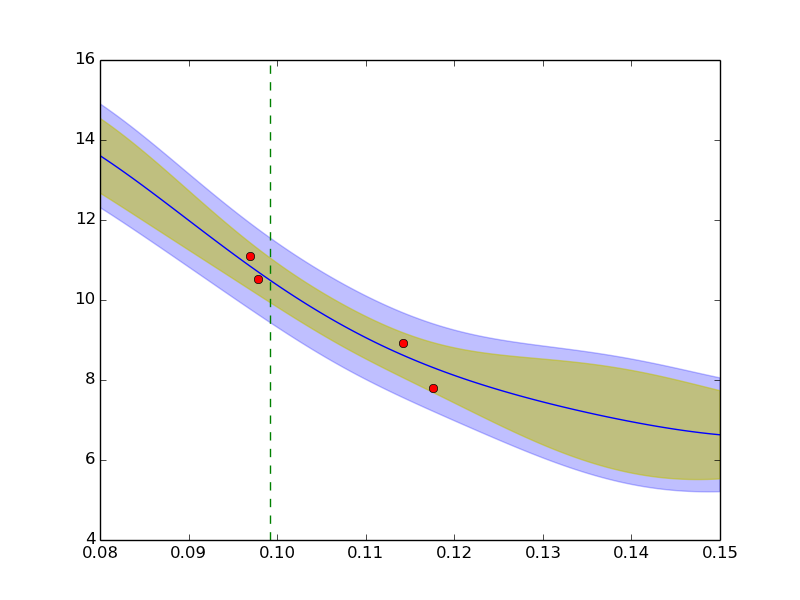}
\hskip -0.1in
\includegraphics[width=0.33\columnwidth]{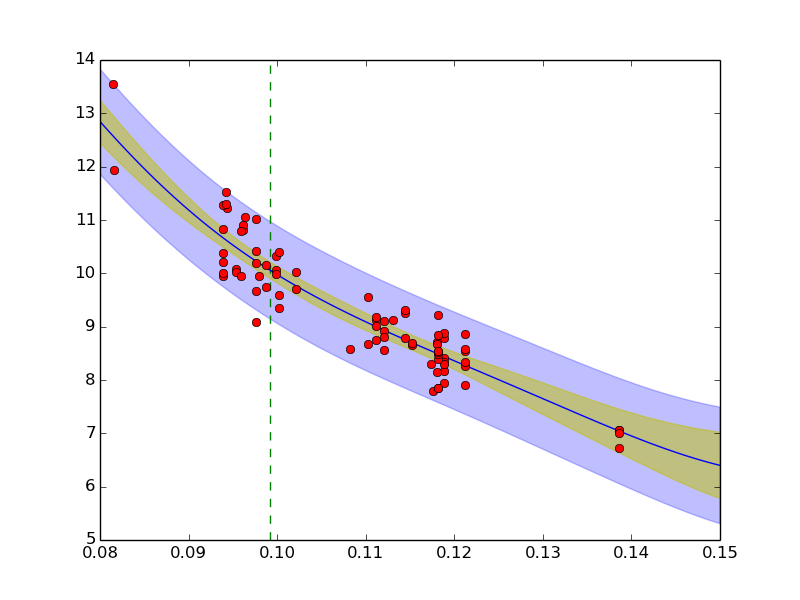}
\hskip -0.1in
\includegraphics[width=0.33\columnwidth]{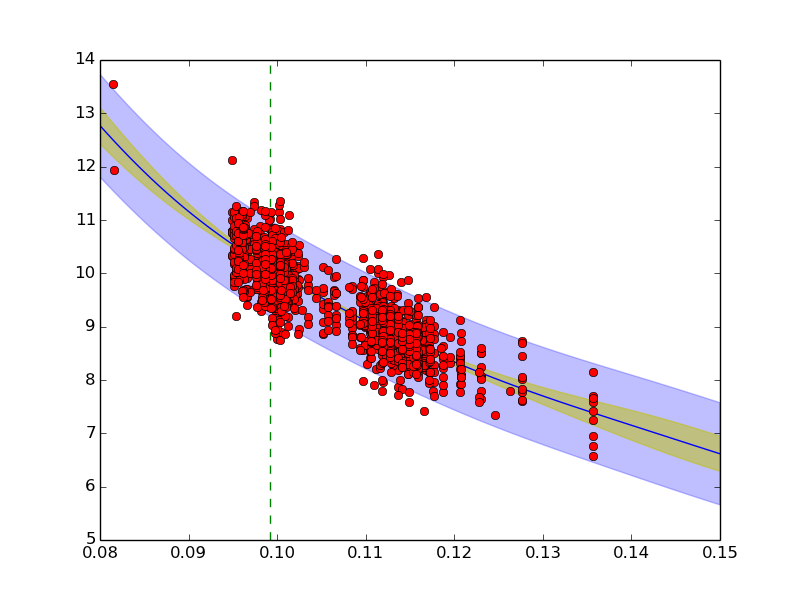}
\caption{\small{The final GPs from Figure~\ref{fig:toycompareposteriors}, zoomed in near their posterior mode (from left to right, $\xi = \{0.4,0.2,0.05\}$).  The y-axis represents $\x$; note that $\y = 9.42$.  The x-axis represents $\thetav$; note that posterior mode is indicated by the vertical  line. The model uncertainty is indicated in yellow (interior shading) within the total uncertainty (2 stds for both).  }}
\label{fig:gpzoom}
\end{center}
\vskip -0.2in
\end{figure}

The number of simulations used by the algorithms is shown in Table~\ref{table-exponential}. The results for Kernel-ABC can be deceptive.  Although Kernel-ABC  performs quite well on this toy example (its over-dispersed results for small $\eps$ can be corrected using the pseudo-marginal sampler at the cost of a higher rejection rate), its performance should be taken with a grain of salt: Kernel-ABC has an extremely high rejection rate ($>0.99$ for $\eps=0.01$) and therefore mixes very slowly (as noted in Figure~\ref{fig:toycompareposteriors}, the sampler was run longer to provide a smoother histogram).  For this reason, though not pursued in this paper, a more useful measure of performance may be the rate of independent samples per simulation call.   For the ASL-ABC the results are good even for large $\xi$.  Even though the distributions are more accurate when $S_0 = 100$ (not shown in figure), for $S_0 = 5$, using an adaptive number of simulations $S$ produces many fewer simulations than for $S_0=100$.    For GPS-ABC the results show a significant advantage over ASL-ABC and Kernel-ABC.  For the most accurate run, where $\xi = 0.05$, only $1297$ simulations were required (compared to over $500K$ for ASL-ABC).  With $\xi=0.2$, nearly the same accuracy is achieved with only $184$ simulations (compared to $100K$ for ASL-ABC). The final GP models are shown in Figure~\ref{fig:gpzoom}, zoomed in to the region near the posterior mode (shown as dashed vertical).  The mean and total uncertainty (2 stds) are shown with interior (yellow) and exterior (blue) shading, respectively; training points are shown as (red) circles.  Notice how the GP model uncertainty contracts as $\xi$ decreases; this an effect produced by adding training points at uncertain locations.
%

\subsection{Chaotic Ecological Systems}\label{experiments}

\begin{figure}[t]
\begin{center}
  \vskip -0.1in
\centerline{\includegraphics[width=0.8\columnwidth]{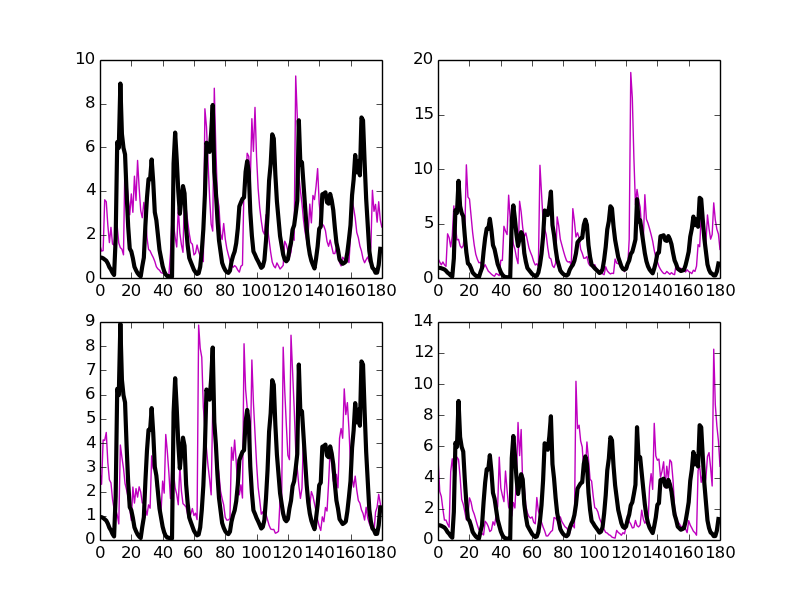}}
\vskip -0.1in
\caption{\small{The blowfly population time-series.  The observations are in bold (black) lines, replicated $4$ times along with generated time-series (light/magenta lines) based on $\thetav$ samples from our algorithms.  Note the observations are the same in each plot; the scale changes due to the generated time-series.}}
\label{fig:blowflyexamples}
\end{center}
\end{figure}

Adult blowfly populations exhibit dynamic behavior for which several competing population models exist.  In this experiment, we use observational data and a simulation model from Wood \cite{wood2010statistical}, based on their improvement upon previous population dynamics' theory.  Population dynamics are modeled using (discretized) differential equations that can produce chaotic behavior for some parameter settings.  We use model 4 from \cite{wood2010statistical} as our simulator; the generated time-series are then processed into $4$ statistics explained below.  An example blowfly replicate series is shown in Figure~\ref{fig:blowflyexamples}, along with several  times-series generated from $\p(\thetav|\y)$.

The population dynamics' equation generates  $N_1, \ldots, N_T$ using the following update rule:
\begin{equation}
N_{t+1} = P N_{t-\tau} \exp(-N_{t-\tau}/N_0) e_t + N_t \exp(\partial \epsilon_t) \nonumber
\end{equation}
where $e_t \sim  \mathcal{G}( 1/{\sigma_p^2},1/{\sigma_p^2})$ and $\epsilon_t 
 \sim  \mathcal{G}( 1/{\sigma_d^2},1/{\sigma_d^2})$  
are sources of noise, and $\tau$ is an integer (not to be confused with the $\tau$ used as the MH acceptance threshold in our algorithms).  In total, there are 6 parameters $\theta = \{ P, \partial, N_0, \sigma_d, \sigma_p, \tau\}$.  See \cite{wood2010statistical} for further details about the significance of the parameters.

We model the likelihood of $4$ statistics: 1) the mean of $\{N_t\}$, 2) the mean minus the median, 3) the maximal peaks in the time-series with a smooth detection threshold, and 4) the log of the maximum of $\{N_t\}$.  Our first two statistics are used by \cite{wood2010statistical}, but they have 21 other statistics we do not use. The reasons we do not choose all $23$ are 1) we did not wish to inaccurately replicate the statistics using the code from \cite{wood2010statistical}, 2) our goal is sampling given a set of statistics (not finding the optimal statistics), and 3) time-series simulated with parameters based on our statistics are a reasonable representation of the observations (see Figure~\ref{fig:blowflyexamples}).  

Like \cite{wood2010statistical} we model the log distribution of $\thetav$, with the following broad (in the non-log space) $\pi(\log\thetav)$: $\log P\sim \mathcal{N}(2,2^2)$, $\log \partial \sim \mathcal{N}(-1.8,0.4^2)$, $\log N_0\sim\mathcal{N}(6,0.5^2)$, $\log \sigma_d \sim \mathcal{N}(-0.75,1^2)$, $\log \sigma_p \sim \mathcal{N}(-0.5,1^2)$, $\log \tau \sim\mathcal{N}(2.7,0.1^2)$.
Time-series generated with parameters from this prior distribution produce extremely varied results, some are chaotic, some are degenerate, etc.  Modeling this simulator is {\em very} challenging.

\begin{figure}[ht]
\begin{center}
\vskip -0.1in
\includegraphics[width=0.8\columnwidth]{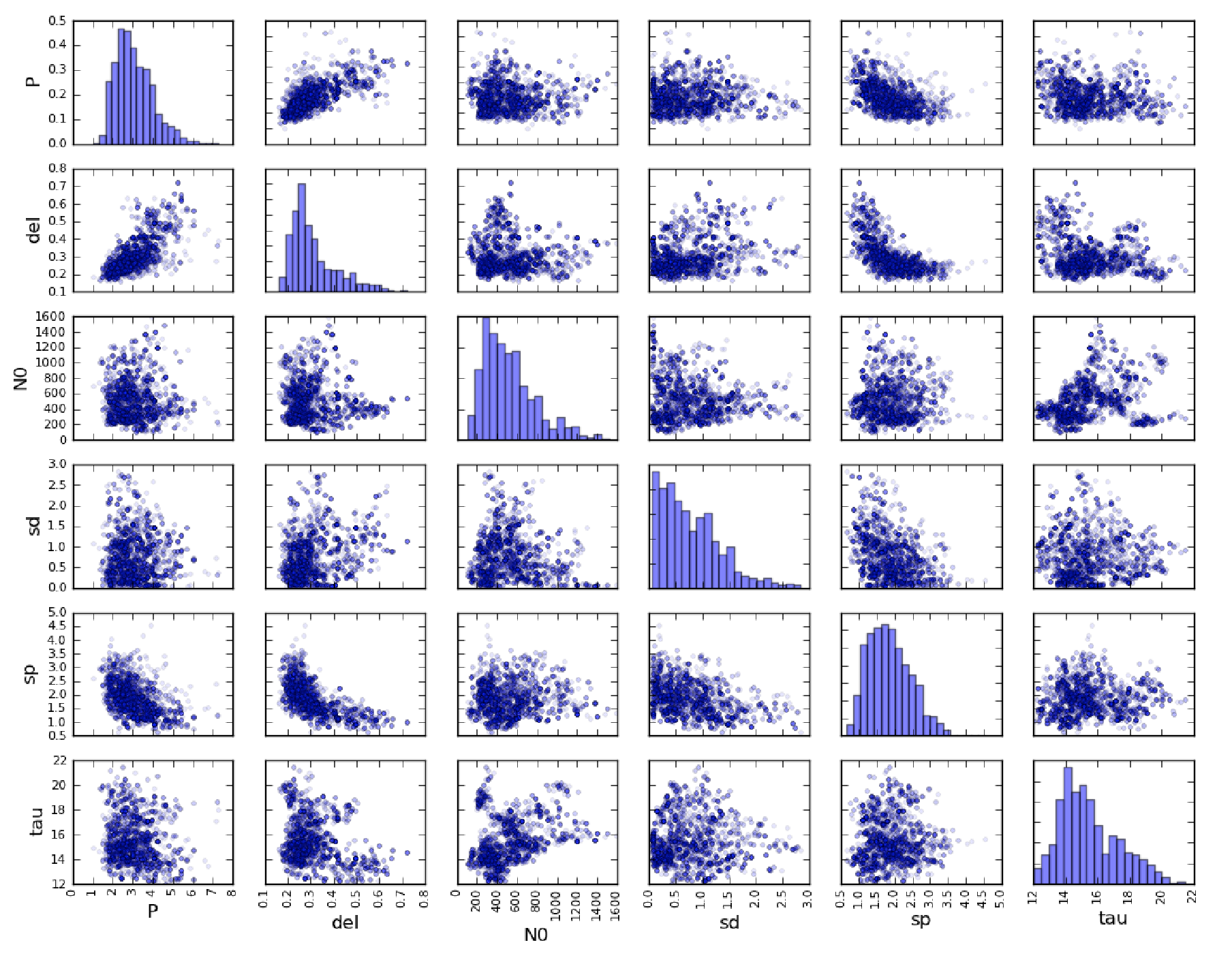}
\vskip -0.1in
\caption{\small{Posterior of $\thetav$ using ASL-ABC on the blowfly simulator, with $\xi = 0.05$ and full-covariance estimators, showing the last 3K samples.  Required approximately $1.12$ million simulation calls.}}
\label{fig:blowflyscatter_synth}
\end{center}
\end{figure} 

\begin{figure}[ht]
\begin{center}
\vskip -0.1in
\includegraphics[width=0.8\columnwidth]{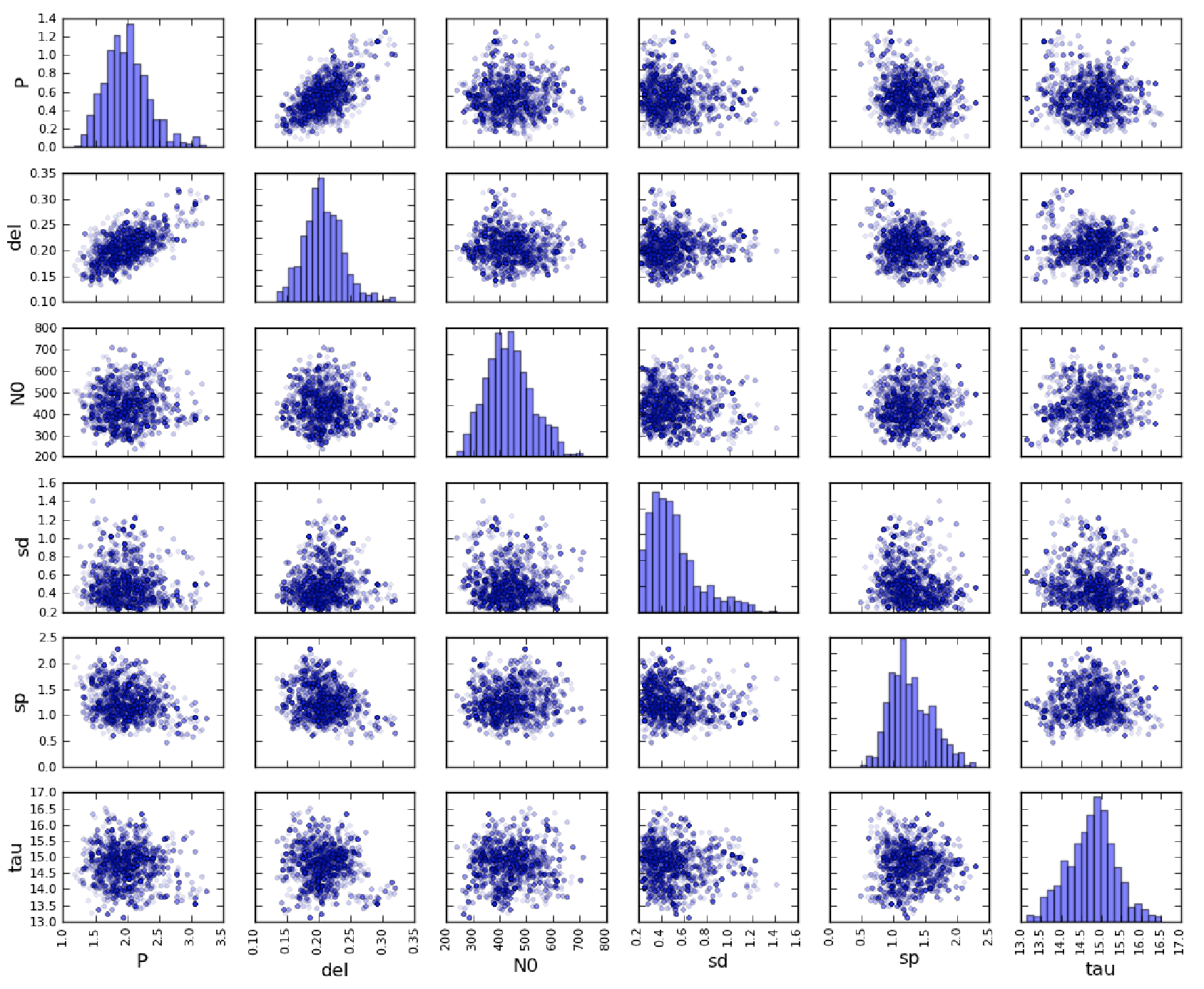}
\vskip -0.1in
\caption{\small{Posterior of $\thetav$ using GPS-ABC  on the blowfly simulator.    Required approximately $3$K simulation calls.}}
\label{fig:blowflyscatter_gps}
\end{center}
\end{figure}

We compared the results from using ASL-ABC with $S_0=100$, $\xi=0.05$, and $\epsilon=0$ against GPS-ABC with $S_0=1000$, $\xi=0.3$, and $\epsilon=0$. (Recall that generating $S_0$  samples for GPS-ABC occurs only once at initialization, while for ASL-ABC twice at every MH-step.)  We set $\xi=0.3$ for GPS-ABC to control the number of simulations at each time step.  For both algorithms the proposal distribution was a zero-mean Gaussian with standard deviation equal to $1/5$ the prior's value, with only a single dimension proposed at each MH step.  

\begin{figure}[th]
  \begin{center}
  \includegraphics[width=\columnwidth]{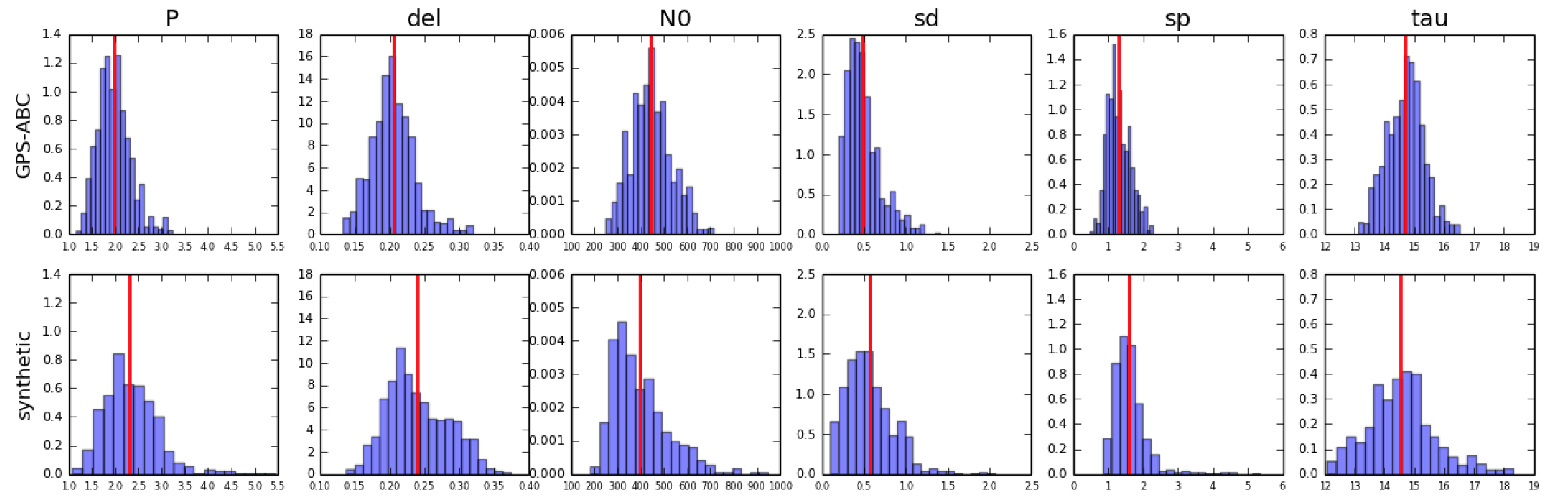} 
  \caption{\small{Blowfly results (both diagonal).  The posterior samples of $\thetav$ for GPS-ABC ({\bf top}) and ASL-ABC ({\bf bottom}); both x-axes are at the same scale.  Mean posteriors are shown as vertical lines. When both methods use independent output models, the results are very similar; compare this to the full-covariance results for ASL-ABC in Figure~\ref{fig:blowflyscatter_synth} where the posteriors have heavier tails.}}  
  \label{fig:blowfly}
  \end{center}
  \end{figure}

Results are shown in the form of scatter plots for $p(\thetav | \y)$ in Figures~\ref{fig:blowflyscatter_synth} and \ref{fig:blowflyscatter_gps}.  Note we are displaying the parameters in $\thetav$-space, not log-$\thetav$.  Results for the ASL-ABC show interesting non-linear relationships between parameters.  On close inspection of the results it can be seen that the posterior modes for the two models are very similar.  However, the variance of the ASL-ABC is larger than for GPS-ABC.  This is most likely due to the independence assumption made by GPS-ABC over the $J$ output statistics.  We re-ran ASL-ABC with covariance estimators forced to be diagonal, and the results were very similar to GPS-ABC (see Figure~\ref{fig:blowfly}).  On the one hand the difference between full and diagonal covariance can be interpreted as overconfidence in the posterior, and on the other it can can be viewed as a robust approximation to the full posterior, because it is very difficult to determine, for the ASL-ABC, the degree to which the covariance estimator is influenced by outliers (and overfits, in a sense).  By enforcing diagonal likelihoods for the ASL-ABC, similar variance in the posterior distributions are found.  Despite the potential lack of a fully informative posterior, GPS-ABC results were achieved with only a few thousand simulation calls, whereas ASL-ABC used well over a million calls.  If the cost per simulation is very expensive, the trade-off of accuracy versus computation leans heavily towards GPS-ABC.
%

\subsection{Feedback Wright-Fisher model}

The Wright-Fisher model is a standard stochastic model for reproduction in population genetics.  In two experiments, we study the potential of our methods for model selection and comparison (rather than computational savings) using the ASL-ABC (similar results can be obtained using GPS-ABC).  
We apply our methods to the model in \cite{ZhuWJL12}, where there are three types of cell populations: stem cells (SC), transit-amplifying cells (TAC), and terminal differentiated cells (TDC).  A simulation begins with an initial cell population that is updated using stochastic equations having parameters $\thetav = \{ N,k,p_1\}$.  The true parameter setting is $\thetav^{\star} = \{ 2000, 3, 0.1\}$.  A simulation consists of running the equations for $500$ time-steps, repeated $10$ times.  Statistics are the mean and log-variance of these population sizes.  Due to space considerations, only the results for the {\em mean size} are shown in the figures. 

For this experiment we have slightly changed notation for the statistics.  Here we call $\y^{\star}$ the {\em true} statistics (whereas previously we simply referred to these as the observed statistics $\y$) and $p(\y | \y^{\star})$ as the {\em posterior predictive distribution} defined as
\begin{equation}
  p(\y | \y^{\star}) = \int p(\y | \thetav ) p( \thetav | \y^{\star} ) d \thetav
\end{equation} 
where $p( \thetav | \y^{\star} )$ is the target distribution of our MCMC sampler ($\pi(\thetav | \y)$, previously).  Therefore the true statistics $\y^{\star}$ is the result of a single (arbitrarily chosen) simulation using $\thetav^{\star}$.  We set broad priors over $\thetav$.  

\begin{figure}[ht]

\centering
\vskip -0.1in
\begin{subfigure}[b]{\columnwidth}
\includegraphics[width=\columnwidth]{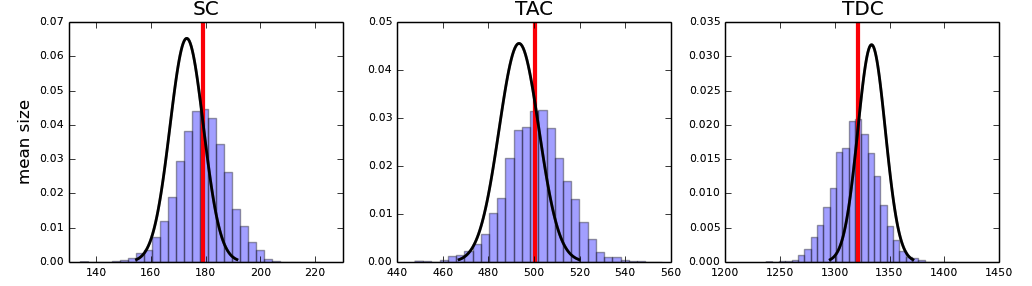} 
\vskip -0.1in
\caption{ \small{The distribution $p(\y | \y^{\star})$ when inferring $\thetav = \{N, k, p_1 \}$. }}
\end{subfigure}
\vskip 0.2in
\begin{subfigure}[b]{\columnwidth}
\includegraphics[width=\columnwidth]{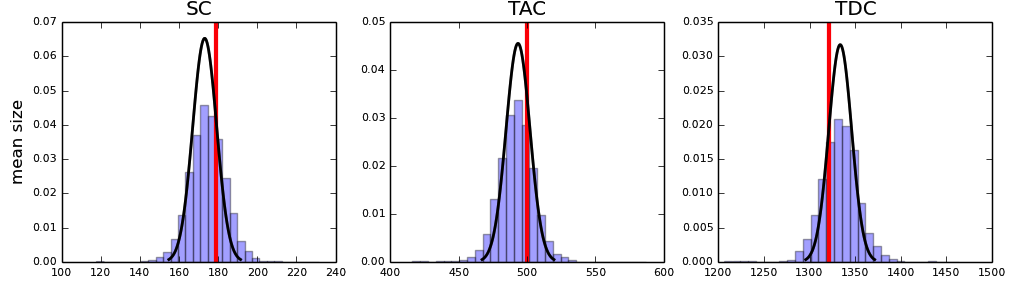} 
\vskip -0.05in
\caption{\small{ The distribution $p(\y | \y^{\star})$ when inferring $\thetav = \{N, p_1 \}$ and fixing $k$ to its true value from $\thetav^{\star}$.}}
\end{subfigure}
\vskip 0.2in
\begin{subfigure}[b]{\columnwidth}
\includegraphics[width=\columnwidth]{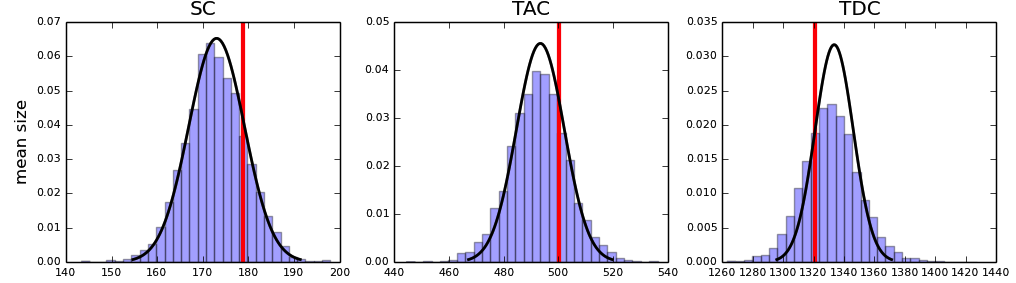} 
\vskip -0.05in
\caption{ \small{The distribution $p(\y | \y^{\star})$ when inferring $\thetav = \{N \}$ and fixing $k$ and $p_1$ fixed to their true values from $\thetav^{\star}$. }}
\end{subfigure}
\vskip -0.05in
\caption{\small{Wright-Fisher experiment 1 results.  In each plot are the posterior predictive distributions $p(\y | \y^{\star})$ for population {\em mean sizes}.  The number of parameters $\thetav$ inferred in the MCMC run changes for each row of the figure.  The distribution $p(\y | \thetav^{\star})$ is shown as a solid curve in each plot.  The vertical lines indicate $\y^{\star}$.  See text for discussion.  }}
\label{fig:wf}

\end{figure}

In Figure~\ref{fig:wf}, we show results of experiment 1.  In this experiment, the number of parameters $\thetav$ varies, allowing us to compare $p(\y | \y^{\star})$ as the number of parameters that are inferred changes.  From top to bottom, we infer $\thetav = \{N, k, p_1 \}$, $\thetav = \{N, p_1 \}$, and  $\thetav = \{N \}$, respectively, and set any missing parameters to their corresponding value in $\thetav^{\star}$.  The solid curve is $p(\y | \thetav^{\star})$ (the distribution of statistics generated at $\thetav^{\star}$).  The true $\y^{\star}$ is a draw from this distribution (shown as vertical line).  When the algorithm infers all three parameters, the posterior predictive distribution is centered at $\y^{\star}$, whereas when it infers only a single parameter, it is shifted towards $p(\y | \thetav^{\star})$.  If we inspect $p(\y | \y^{\star})$, we can see that the value for $\y^{\star}$ is the same for all three rows.  For the first row, this can only be achieved by having a larger variance in the posterior predictive distribution; for the last row, this is achieved by decreasing the variance.

\begin{figure}[ht]

\centering
\vskip -0.1in
\begin{subfigure}[b]{\columnwidth}
\includegraphics[width=\columnwidth]{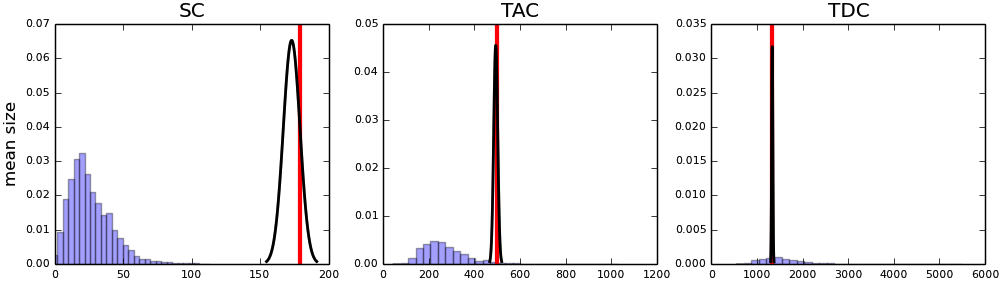} 
\vskip -0.01in
\caption{ \small{The distribution $p(\y | \y^{\star})$ when inferring $\thetav = \{N, p_1 \}$ and fixing $k$ to a ``bad'' value.  The vertical line indicates $\y^{\star}$.  The solid curve is $p(\y | \thetav^{\star})$.}}
\end{subfigure}
\begin{subfigure}[b]{0.83\columnwidth}
\includegraphics[width=\columnwidth]{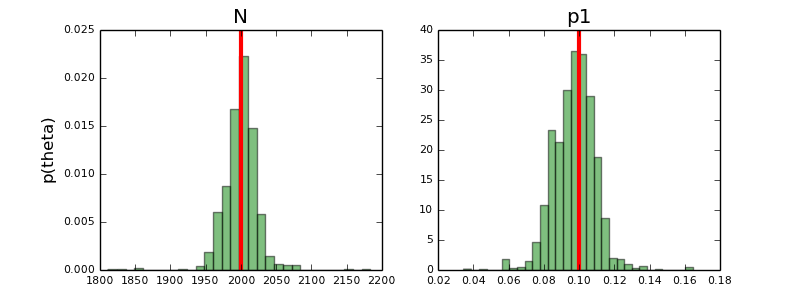} 
\vskip -0.01in
\caption{ \small{ The distribution $p(\thetav | \y^{\star})$  when inferring $\thetav = \{N, p_1 \}$ and fixing $k$ to a ``good'' value. The vertical line indicates $\thetav^{\star}$.}}
\end{subfigure} \hskip -0.02in
\begin{subfigure}[b]{0.8\columnwidth}
\includegraphics[width=\columnwidth]{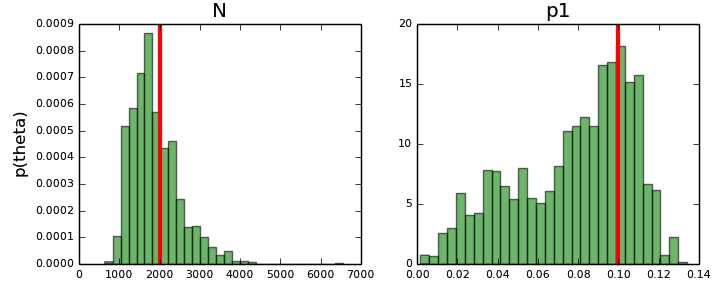} 
\vskip -0.01in
\caption{ \small{The distribution $p(\thetav | \y^{\star})$  when inferring $\thetav = \{N, p_1 \}$ and fixing $k$ to a ``bad'' value.  The vertical line indicates $\thetav^{\star}$.} }
\end{subfigure}

\vskip -0.1in
\caption{\small{ Wright-Fisher experiment 2 results.  See text for discussion.}}
\label{fig:wf2}

\end{figure}

In experiment 2, we compare the result for the middle row in Figure~\ref{fig:wf} where $k$ is set to the true or ``good" setting ($k=3$), with $k$ set to a ``bad" ($k=6$).  In Figure~\ref{fig:wf2} top row, we show posterior predictive distribution $p(\y | \y^{\star})$ for the latter case; notice that these posterior predictive distributions show evidence of large bias and large variance.  The bottom row compares the posterior distributions $p( \thetav | \y^{\star} )$ for both cases.  Here the bias is less pronounced, but the variance remains large.   

The results of these two experiments demonstrate the potential of our approach for model checking, selection, and improvement described by Gelman et al. \cite{gelman} (chapter 6).  This type of analysis  comes naturally for our algorithms for two reasons:  the ABC statistics $\y$ are natural choices for Gelman et al.'s model checking statistics and these statistics are pre-computed as part of the MH-step of our algorithms.  We can now answer questions like: what is the likelihood of a set of observations $\y^{\star}$ using simulator $a$ versus simulator $b$?
%
        
\section{Discussion and Future Work}\label{discussion}
We have presented a promising framework for performing inference in expensive, simulation-based models.  Our algorithms improve current state-of-the-art ABC methods in that they require many fewer calls to the simulator.  

Using GPs for surrogate modeling has an appealing elegance; as nonparametric Bayesian models, they naturally incorporate both model and pseudo-data uncertainty into the MCMC algorithms.  However, there are several technical issues and modeling limitations with GPs used for surrogate modeling. Heteroskedatic noise is more likely the norm than the exception for complicated simulators.  The blowfly simulator is a prime example of this.  Improvements to our GPs may be achieved using an input-dependent noise model \cite{goldberg:1998,kersting:2007}, where the noise is an additional independent Gaussian process.  Another limitation of our GP model is the output independence assumption. A more realistic assumption is a full covariance Gaussian process such as the convolution processes of \cite{higdon:2002,boyle:2005,alvarez:2011}. The other major difficulty is setting the GP hyper-parameters and updating them during the ABC run. 

Our GPS-ABC uses a random-walk proposal distribution which is inefficient for exploring the target distribution.  Using GPs offers the opportunity to use other techniques to improve the mixing (and in turn computational cost).   For example, in \cite{rasmussen:2003} a Hamiltonian Monte Carlo run on the GP surface is used to generate {\em independent} proposals. If applied to ABC, their algorithm would require the equivalent of a full simulation at the proposed location, whereas if we incorporated a similar technique, we would then test the GP uncertainty to determine if a simulation was required.  

There has been a recent surge in interest in Bayesian optimization using Gaussian process (GP-BO) surrogates of the objective surface \cite{brochu:2010,snoek:2012}. GP-BO is often applied to problems where simulation or sometimes user feedback guides the surrogate's construction.  What is most interesting about GP-BO is its use of model uncertainty to {\em actively} determine the next simulation location implemented through acquisition functions. These ideas can probably be generalized to our GPS-ABC algorithm to further reduce simulation costs, while at the same time maintaining control of the MCMC error. The authors of \cite{snoek:2012} also develop distributed algorithms for optimization that may serve as inspiration for parallel ABC sampling algorithms.

\newpage
\bibsep0pt
{\small
\bibliographystyle{icml2014}
\bibliography{skabc}

\begin{thebibliography}{30}
\providecommand{\natexlab}[1]{#1}
\providecommand{\url}[1]{\texttt{#1}}
\expandafter\ifx\csname urlstyle\endcsname\relax
  \providecommand{\doi}[1]{doi: #1}\else
  \providecommand{\doi}{doi: \begingroup \urlstyle{rm}\Url}\fi

\bibitem[{\'A}lvarez \& Lawrence(2011){\'A}lvarez and Lawrence]{alvarez:2011}
{\'A}lvarez, M. and Lawrence, L.D.
\newblock Computationally efficient convolved multiple output gaussian
  processes.
\newblock \emph{Journal of Machine Learning Research}, 12:\penalty0 1425--1466,
  2011.

\bibitem[Andrieu \& Roberts(2009)Andrieu and Roberts]{andrieu2009pseudo}
Andrieu, C. and Roberts, G.
\newblock The pseudo-marginal approach for efficient monte carlo computations.
\newblock \emph{The Annals of Statistics}, 37\penalty0 (2):\penalty0 697--725,
  2009.

\bibitem[Bazin et~al.(2010)Bazin, Dawson, and Beaumont]{bazin2010likelihood}
Bazin, Eric, Dawson, Kevin~J, and Beaumont, Mark~A.
\newblock Likelihood-free inference of population structure and local
  adaptation in a bayesian hierarchical model.
\newblock \emph{Genetics}, 185\penalty0 (2):\penalty0 587--602, 2010.

\bibitem[Beaumont et~al.(2002)Beaumont, Zhang, and
  Balding]{beaumont2002approximate}
Beaumont, Mark~A, Zhang, Wenyang, and Balding, David~J.
\newblock Approximate bayesian computation in population genetics.
\newblock \emph{Genetics}, 162\penalty0 (4):\penalty0 2025--2035, 2002.

\bibitem[Beaumont et~al.(2009)Beaumont, Cornuet, Marin, and
  Robert]{beaumont2009adaptive}
Beaumont, Mark~A, Cornuet, Jean-Marie, Marin, Jean-Michel, and Robert,
  Christian~P.
\newblock Adaptive approximate bayesian computation.
\newblock \emph{Biometrika}, 96\penalty0 (4):\penalty0 983--990, 2009.

\bibitem[Bilionis et~al.(2013)Bilionis, Zabaras, Konomi, and Lin]{bilionis2013}
Bilionis, I., Zabaras, N., Konomi, B.A., and Lin, G.
\newblock Multi-output separable gaussian process: Towards an efficient fully
  bayesian paradigm for uncertainty quantification.
\newblock \emph{Journal of Computational Physics}, 241:\penalty0 212--239,
  2013.

\bibitem[Boyle \& Frean(2005)Boyle and Frean]{boyle:2005}
Boyle, P. and Frean, M.
\newblock Dependent gaussian processes.
\newblock \emph{Advances in Neural Information Processing Systems 17}, 2005.

\bibitem[Brochu et~al.(2010)Brochu, Cora, , and de~Freitas]{brochu:2010}
Brochu, E., Cora, Vlad~M., , and de~Freitas, Nando.
\newblock A tutorial on bayesian optimization of expensive cost functions, with
  application to active user modeling and hierarchical reinforcement learning.
\newblock Technical report, UBC Technical Report, 2010.

\bibitem[Ceperley \& Dewing(1999)Ceperley and Dewing]{ceperley:1999}
Ceperley, D.M. and Dewing, M.
\newblock The penalty method for random walks with uncertain energies.
\newblock \emph{Journal of Chemical Physics}, 110\penalty0 (20):\penalty0
  9812--9820, 1999.

\bibitem[Gelman et~al.(2004)Gelman, Carlin, Stern, and Rubin]{gelman}
Gelman, A., Carlin, J.B., Stern, H.S., and Rubin, D.B.
\newblock \emph{Bayesian Data Analysis}.
\newblock Chapman and Hall/CRC, New York, second edition, 2004.

\bibitem[Goldberg et~al.(1998)Goldberg, Williams, and Bishop]{goldberg:1998}
Goldberg, P.W., Williams, C.K.I., and Bishop, C.M.
\newblock Regression with input-dependent noise: A gaussian process treatment.
\newblock \emph{Advances in Neural Information Processing Systems 10}, 1998.

\bibitem[Higdon(2002)]{higdon:2002}
Higdon, D.
\newblock Space and space-time modeling using process convolutions.
\newblock In Anderson, C.W., Barnett, V., Chatwin, P.C, and Abdel, E.-S.H.
  (eds.), \emph{Quantitative Methods for Current Environmental Issues}, pp.\
  37--56. Springer London, 2002.
\newblock ISBN 978-1-4471-1171-9.

\bibitem[Kersting et~al.(2007)Kersting, Plagemann, Pfaff, and
  Burgard]{kersting:2007}
Kersting, Kristian, Plagemann, Christian, Pfaff, Patrick, and Burgard, Wolfram.
\newblock Most likely heteroscedastic gaussian process regression.
\newblock In \emph{Proceedings of the 24th international conference on Machine
  learning}, pp.\  393--400. ACM, 2007.

\bibitem[Korattikara et~al.(2014)Korattikara, Chen, and
  Welling]{korrattikara:2014}
Korattikara, A., Chen, Y., and Welling, M.
\newblock Austerity in mcmc land: Cutting the metropolis-hastings budget.
\newblock \emph{ICML}, 2014.
\newblock To appear.

\bibitem[Marin et~al.(2012)Marin, Pudlo, Robert, and Ryder]{marin:2012}
Marin, J.-M., Pudlo, P., Robert, C.P., and Ryder, R.J.
\newblock Approximate bayesian computational methods.
\newblock \emph{Statistics and Computing}, 22:\penalty0 1167--1180, 2012.

\bibitem[Nicholls et~al.(2012)Nicholls, Fox, and Watt]{nicholls:2012}
Nicholls, G.K., Fox, C., and Watt, A.M.
\newblock Coupled mcmc with a randomized acceptance probability.
\newblock Technical report, arXiv:1205.6857v1, 2012.

\bibitem[Rasmussen(2003)]{rasmussen:2003}
Rasmussen, C.E.
\newblock Gaussian processes to speed up hybrid monte carlo for expensive
  bayesian integrals.
\newblock \emph{Bayesian Statistics}, 7:\penalty0 651--659, 2003.

\bibitem[Ratmann et~al.(2007)Ratmann, Jorgensen, Hinkley, Stumpf, Richardson,
  and Wiuf]{ratmann2007using}
Ratmann, O., Jorgensen, O., Hinkley, T., Stumpf, M., Richardson, S., and Wiuf,
  C.
\newblock Using likelihood-free inference to compare evolutionary dynamics of
  the protein networks of h. pylori and p. falciparum.
\newblock \emph{PLoS Computational Biology}, 3\penalty0 (11):\penalty0 e230,
  2007.

\bibitem[Roberts \& Rosenthal(2007)Roberts and Rosenthal]{roberts:2007}
Roberts, G.O. and Rosenthal, J.S.
\newblock Coupling and ergodicity of adaptive mcmc.
\newblock \emph{J. Appl. Prob.}, 44:\penalty0 458--475, 2007.

\bibitem[Schafer \& Freeman(2012)Schafer and Freeman]{schafer2012likelihood}
Schafer, C.M. and Freeman, P.E.
\newblock Likelihood-free inference in cosmology: Potential for the estimation
  of luminosity functions.
\newblock In \emph{Statistical Challenges in Modern Astronomy V}, pp.\  3--19.
  Springer, 2012.

\bibitem[Sisson et~al.(2007)Sisson, Fan, and Tanaka]{sisson2007sequential}
Sisson, SA, Fan, Y, and Tanaka, Mark~M.
\newblock Sequential monte carlo without likelihoods.
\newblock \emph{Proceedings of the National Academy of Sciences}, 104\penalty0
  (6):\penalty0 1760, 2007.

\bibitem[Sisson et~al.(2010)Sisson, Peters, Briers, and Fan]{sisson:2008}
Sisson, S.A., Peters, G.W., Briers, M., and Fan, Y.
\newblock A note on target distribution ambiguity of likelihood-free samplers.
\newblock \emph{Arxiv preprint arXiv:1005.5201v1 [stat.CO]}, 2010.

\bibitem[Sisson \& Fan(2010)Sisson and Fan]{sisson:2010}
Sisson, Scott~A and Fan, Yanan.
\newblock Likelihood-free markov chain monte carlo.
\newblock \emph{Arxiv preprint arXiv:1001.2058}, 2010.

\bibitem[Snoek et~al.(2012)Snoek, Larochelle, and Adams]{snoek:2012}
Snoek, Jasper, Larochelle, Hugo, and Adams, Ryan~Prescott.
\newblock Practical bayesian optimization of machine learning algorithms.
\newblock \emph{Advances in Neural Information Processing Systems 25}, 2012.

\bibitem[Tavare et~al.(1997)Tavare, Balding, Griffiths, and
  Donnelly]{tavare1997inferring}
Tavare, S., Balding, D.J., Griffiths, R.C., and Donnelly, P.
\newblock Inferring coalescence times from dna sequence data.
\newblock \emph{Genetics}, 145\penalty0 (2):\penalty0 505--518, 1997.

\bibitem[Toni et~al.(2009)Toni, Welch, Strelkowa, Ipsen, and
  Stumpf]{toni2009approximate}
Toni, Tina, Welch, David, Strelkowa, Natalja, Ipsen, Andreas, and Stumpf,
  Michael~PH.
\newblock Approximate bayesian computation scheme for parameter inference and
  model selection in dynamical systems.
\newblock \emph{Journal of the Royal Society Interface}, 6\penalty0
  (31):\penalty0 187--202, 2009.

\bibitem[Turner \& Van~Zandt(2012)Turner and Van~Zandt]{turner2012tutorial}
Turner, Brandon~M and Van~Zandt, Trisha.
\newblock A tutorial on approximate bayesian computation.
\newblock \emph{Journal of Mathematical Psychology}, 56\penalty0 (2):\penalty0
  69--85, 2012.

\bibitem[Wood(2010)]{wood2010statistical}
Wood, Simon~N.
\newblock Statistical inference for noisy nonlinear ecological dynamic systems.
\newblock \emph{Nature}, 466\penalty0 (7310):\penalty0 1102--1104, 2010.

\bibitem[Xue \& Titterington(2011)Xue and Titterington]{xue:2011}
Xue, Jing-Hao and Titterington, D.~Michael.
\newblock The {\em p}-folded cumulative distribution function and the mean
  absolute deviation from the {\em p}-quantile.
\newblock \emph{Statistics and Probability Letters}, 81:\penalty0 1179--1182,
  2011.

\bibitem[Zhu et~al.(2012)Zhu, Welling, Jin, and Lowengrub]{ZhuWJL12}
Zhu, X., Welling, M., Jin, F., and Lowengrub, J.S.
\newblock Predicting simulation parameters of biological systems using a
  gaussian process model.
\newblock \emph{Statistical Analysis and Data Mining}, 5\penalty0 (6):\penalty0
  509--522, 2012.

\end{thebibliography}
}
\end{document}